\documentclass[preprint,12pt]{elsarticle}

\usepackage{amssymb}
\usepackage{amsmath}
\usepackage{bm}
\usepackage{algorithmic}
\usepackage{algorithm}
\usepackage{subfig}
\usepackage{booktabs}
\usepackage{multirow}
\usepackage{newtxtext}
\usepackage{makecell}
\usepackage{url}
\usepackage[colorlinks=false]{hyperref}
\usepackage{geometry}
\usepackage{color}

\begin{document}

\begin{frontmatter}



\title{Expected Coordinate Improvement for High-Dimensional Bayesian Optimization}


\author{Dawei~Zhan\corref{cor}}
\ead{zhandawei@swjtu.edu.cn}
\cortext[cor]{Corresponding author}

\affiliation{organization={School of Computing and Artificial Intelligence, Southwest Jiaotong University},
            city={Chengdu},
            country={China}}

\begin{abstract}
Bayesian optimization (BO) algorithm is very popular for solving low-dimensional expensive optimization problems. Extending Bayesian optimization to high dimension is a meaningful but challenging task. One of the major challenges is that it is difficult to find good infill solutions as the acquisition functions are also high-dimensional. In this work, we propose the expected coordinate improvement (ECI) criterion for high-dimensional Bayesian optimization. The proposed ECI criterion measures the potential improvement we can get by moving the current best solution along one coordinate.  The proposed approach selects the coordinate with the highest ECI value to refine in each iteration and covers all the coordinates gradually by iterating over the coordinates.  The greatest advantage of the proposed ECI-BO (expected coordinate improvement based Bayesian optimization) algorithm over the standard BO  algorithm is that the infill selection problem of the proposed algorithm is always a one-dimensional problem thus can be easily solved. Numerical experiments show that the proposed algorithm can achieve significantly better results than the standard BO algorithm and competitive results when compared with five high-dimensional BOs and six surrogate-assisted evolutionary algorithms. This work provides a simple but efficient approach for high-dimensional Bayesian optimization.  A Matlab implementation of our ECI-BO is available at  \url{https://github.com/zhandawei/Expected_Coordinate_Improvement}.

\end{abstract}



\begin{keyword}
Bayesian optimization \sep expensive optimization \sep  high-dimensional optimization \sep expected improvement.

\end{keyword}

\end{frontmatter}


\section{Introduction}
\label{sec_introduction}
Bayesian optimization (BO)~\cite{Mockus_1975,Mockus_1994}, also known as efficient global optimization (EGO)~\cite{Jones_1998} is a class of surrogate-based optimization methods.  It employees a statistical model, often a Gaussian process model~\cite{Rasmussen_2006} to approximate the expensive black-box objective function, and selects new samples for expensive evaluations based on a defined acquisition function, such as the expected improvement, lower confidence bound and probability of improvement~\cite{Jones_2001}.  Due to the use of the Gaussian process model and the informative acquisition function, Bayesian optimization is considered as a sample-efficient approach~\cite{Snoek_2012,Shahriari_2016,Guo_2021}. Bayesian optimization has also been successfully extended for solving multiobjective optimization problems~\cite{Zhang_2010,Yang_2019,Han_2022}, constrained optimization problems~\cite{Basudhar_2012,LiGH_2021} and parallel optimization problems~\cite{Briffoteaux_2020,Chen_2023,Zhan_2023,WangZ_2023}.

Despite these successes, Bayesian optimization is often criticized for its incapability in solving high-dimensional problems~\cite{Shahriari_2016}. As the dimension increases, the number of samples required to cover the design space to maintain the model accuracy increases exponentially. In addition,  it becomes exponentially difficult to find the global optimum of the acquisition function which is often highly nonlinear and highly multi-modal when the dimension of the problem increases.  These issues decrease the sample efficiency of Bayesian optimization on high-dimensional problems.

In the past twenty years, different approaches have been proposed to ease the curse of dimensionality on Bayesian optimization~\cite{Binois_2022}. The variable selection approaches assume that most of the variables have little effect on the objective function and select only the active variables to optimize. The active variables can be identified by the reference distribution variable selection~\cite{Linkletter_2006}, the hierarchical diagonal sampling~\cite{Chen_2012}, or  the indicator-based Bayesian variable selection~\cite{Zhang_2023}.  Instead of applying sensitivity analysis techniques to select variables, one can also select a few variables  based on some probability distribution. The dimension scheduling algorithm ~\cite{Ulmasov_2016} selects a set of variables based on the probability distribution that is proportional to the eigenvalues of the covariance matrix in principle component analysis. While the Dropout approach~\cite{Li_2017} randomly selects a subset of variables to optimize in each iteration. The Monte Calro tree search variable selection (MCTS-VS) approach~\cite{Song_2022} applies the MCTS to divide the variables into important variables and unimportant variables, and only optimizes the important variables in each iteration.

Another idea is to embed a lower-dimensional space into the high-dimensional space, run Bayesian optimization in the subspace, and project the infill solution back to the original high-dimensional space for function evaluation. The random embedding Bayesian optimization (REMBO) algorithm~\cite{Wang_2016} utilizes a random embedding matrix to map the high-dimensional space into a lower one. The embedded subspace is guaranteed to contain an optimum for constraint-free problems. But for box-constrained problems, the projection back to the high-dimensional space may fall outside the bounds. This is often called the non-injectivity issue, and is further addressed in~\cite{Binois_2015,Nayebi_2019,Binois_2020,Letham_2020}. The Bayesian optimization with adaptively expanding subspaces (BAxUS) approach~\cite{Papenmeier_2022} gradually increases the dimension of the subspace along the iterations to improve the probability for the low-dimensional space contraining an optimum. The REMBO~\cite{Wang_2016} algorithm assumes that most of the variables have low effect on the objective function, which is often unrealistic in applications. This assumption is relaxed a little by the sequential random embedding strategy~\cite{Qian_2016}, which allows all the variables being effective but still assumes most of them have bounded effect. Instead of using randomly generated embeddings, the embeddings can also be learned by using supervised learning methods, such as the partial least squares regression~\cite{Bouhlel_2016} and sliced inverse regression~\cite{Zhang_2019}. Nonlinear embedding techniques~\cite{Moriconi_2020b,Griffiths_2020,Siivola_2021,Maus_2022} have also been used for developing high-dimensional BOs, but they are often more computationally expensive than linear embeddings.

Decomposing the high-dimensional space into multiple lower-dimensional subspaces is a popular way to ease the curse of dimensionality.  The additive Gaussian process upper confidence bound (Add-GP-UCB)~\cite{Kandasamy_2015} algorithm assumes the objective function is a summation of several lower-dimensional functions with disjoint variables, performs BO in the subspaces, and combines the solutions of the subspaces for function evaluation.  The assumption about the objective function being additively separable in Add-GP-UCB is often too strong, and is further relaxed to allow projected-additive functions~\cite{Li_2016} and functions with overlapping groups~\cite{Rolland_2018}. When the decomposition is unknown, it is often treated as a hyperparameter and can be learned through maximizing the marginal likelihood~\cite{Kandasamy_2015} or through Gibbs sampling~\cite{Wang_2017}.

Besides variable selection, embedding and decomposition, there are also other approaches for developing high-dimensional BOs. The trust region Bayesian optimization (TuRBO)~\cite{Eriksson_2019} algorithm employees local Gaussian process models and the trust region strategy  to improve the exploitation ability of BO in high-dimensional space. The Bayesian optimization with cylindrical kernels (BOCK)~\cite{Oh_2018} algorithm applies a cylindrical transformation to expand the volume near the center and contract the volume near the boundaries. In line Bayesian optimization, the high-dimensional space is randomly mapped into a one-dimensional line such that the acquisition function can be efficiently solved~\cite{Kirschner_2019}. The line can be selected as a random coordinate, and the resulting approach is called coordinate line Bayesian optimization (CoordinateLineBO)~\cite{Kirschner_2019}. A recent excellent review about high-dimensional BOs can be found in~\cite{Binois_2022}.

Most of current approaches make structural assumptions about the objective function to be optimized. The performance of these algorithms often decreases when the assumptions are not met. Optimizing one coordinate at a time is an appealing approach for solving high-dimensional problems since it can ease the curse of dimensionality when optimizing the acquisition function. Although there exists approach that works with coordinates, such as the CoordinateLineBO~\cite{Kirschner_2019}, there is no rigorous criterion for measuring the amount of improvement along different coordinates.  More importantly, how to determine the order to optimize these coordinates is still an open problem. In this work, we propose the expected coordinate improvement based Bayesian optimization (ECI-BO) to fill these research gaps. The major contributions of this work are listed in the following.
\begin{enumerate}
	\item We propose a new acquisition function named expected coordinate improvement (ECI) to measure the improvement as we move the current best solution along one coordinate. The ECI criterion is derived in closed-form expression.
	\item Based on the ECI criterion, we propose a new approach named expected coordinate improvement based Bayesian optimization (ECI-BO) for solving high-dimensional expensive optimization problems. The proposed ECI-BO calculates the ECI values of each coordinate and determines the order to optimize the coordinates based on the coordinates' maximal ECI values. 
	\item We  study the performance of the proposed ECI-BO and compare it with twelve state-of-the-art high-dimensional approaches using  extensive numerical experiments. The experiment results empirically show the advantages of our approach on solving high-dimensional expensive optimization problems.
\end{enumerate}

The rest of this paper is organized as follows. Section~\ref{section_background} introduces the backgrounds about the Gaussian process model and the Bayesian optimization algorithm.  Section~\ref{section_coordinate_descent} describes the proposed coordinate descent Bayesian optimization algorithm. Section~\ref{section_experiment} presents the corresponding numerical experiments. Conclusions about this paper are given in Section~\ref{section_conclusion}.

\section{Backgrounds}
\label{section_background}
This work considers a single-objective black-box optimization problem
\begin{align}
	\begin{split}
		\text{find:    }  & \bm{x} = [x_1,x_2,\cdots,x_d] \\
		\text{minimize:    }  &    f(\bm{x})\\
		\text{subject to:    }  &    \bm{x} \in \mathcal{X} \subseteq \mathbb{R}^d
	\end{split}
\end{align}
where $d$ is the number of variables, $f$ is the objective function, and $\mathcal{X} = \{\bm{x} \in \mathbb{R}^d: a_i \le x_i \le b_i, ~i=1,2,\cdots,d\}$ is the design space.  The objective function $f$ is assumed to be expensive-to-evaluate and noise-free in this work.

The Bayesian optimization algorithm solves this problem by fitting a statistical model to approximate the black-box objective function, and optimizes an acquisition function to produce a new point for expensive evaluation in each iteration.  The detailed description about Bayesian optimization can be found in~\cite{Jones_1998,Shahriari_2016,Frazier_2018,Wang_2023}. A brief introduction to Bayesian optimization is given in the following.

\subsection{Gaussian Process Model}
The statistical model Bayesian optimization often uses is the Gaussian process (GP) regression model~\cite{Rasmussen_2006}, also known as Kriging model~\cite{Forrester_2008} in engineering design. The GP model puts a prior probability distribution over the objective function and refers the posterior probability distribution after observing some samples~\cite{Rasmussen_2006}. 

The prior distribution GP uses is an infinite-dimensional Gaussian distribution, under which any combination of dimensions is a multivariate Gaussian distribution~\cite{Rasmussen_2006}. The prior is specified by the mean function $m(\bm{x})$ and the covariance function $k(\bm{x},\bm{x}')$. The most common choice of the prior mean function is a constant value. Popular covariance functions are the squared exponential (SE) kernel function and the M\`{a}tern kernel function~\cite{Rasmussen_2006}. The SE kernel can be expressed as
\begin{equation}
	k\big(\bm{x},\bm{x}'\big) = s^2 \exp\left( - \frac{\sum_{i=1}^{d} \big(x_i - x_i'\big)^2    }{2l^2}\right) 
\end{equation}
where $s^2$ is the variance and $l$ is the hyperparameter of the SE kernel. Consider a set of $n$ points $\{\bm{x}^{(1)},\bm{x}^{(2)},\cdots,\bm{x}^{(n)}\}$, the prior distribution on their objective values is
\begin{equation}
		\setlength\arraycolsep{2pt}
		\begin{bmatrix}
			f(\bm{x}^{(1)}) \\
			f(\bm{x}^{(2)}) \\
			\cdots \\
			f(\bm{x}^{(n)})
		\end{bmatrix}  
		\sim \mathcal{N}
		\begin{pmatrix}
			\begin{bmatrix}
				m(\bm{x}^{(1)}) \\
				m(\bm{x}^{(2)}) \\
				\cdots \\
				m(\bm{x}^{(n)})
			\end{bmatrix},
			\begin{bmatrix}
				k(\bm{x}^{(1)},\bm{x}^{(1)}) & \cdots & k(\bm{x}^{(1)},\bm{x}^{(n)}) \\
				k(\bm{x}^{(2)},\bm{x}^{(1)}) & \cdots & k(\bm{x}^{(2)},\bm{x}^{(n)}) \\
				\vdots & \ddots & \vdots  \\
				k(\bm{x}^{(n)},\bm{x}^{(1)}) & \cdots & k(\bm{x}^{(n)},\bm{x}^{(n)}) 
			\end{bmatrix}
		\end{pmatrix}
\end{equation}

Assume we have observed the objective values of the $n$ points $\{f^{(1)},f^{(2)},\cdots,f^{(n)}\}$ and want to predict the objective value of an unobserved point $\bm{x}$. These $n+1$ objective values also follow a joint multivariate Gaussian distribution according to the property of Gaussian process, and the conditional distribution of $f(\bm{x})$ can be computed using Bayes' rule~\cite{Rasmussen_2006}
\begin{equation}
	f(\bm{x})  \mid f^{(1:n)}  \sim \mathcal{N}\left(\mu(\bm{x}), \sigma^2(\bm{x})\right)
\end{equation}
where
\begin{equation}
	\label{GP_mean}
	\mu(\bm{x}) = k\big(\bm{x},\bm{x}^{(1:n)}\big)\bm{K}^{-1}\big(f^{(1:n)}-m(\bm{x}^{(1:n)})\big) + m(\bm{x})
\end{equation}
\begin{equation}
	\label{GP_variance}
	\sigma^2(\bm{x}) = k(\bm{x},\bm{x})  - k\big(\bm{x},\bm{x}^{(1:n)}\big)\bm{K}^{-1}k\big(\bm{x}^{(1:n)},\bm{x}\big).
\end{equation}
In the equations, $f^{(1:n)} = [f^{(1)},f^{(2)},\cdots,f^{(n)}]$, $\bm{x}^{(1:n)} = [\bm{x}^{(1)},\bm{x}^{(2)},\cdots,\bm{x}^{(n)}]$, and $\bm{K}$ is the covariance matrix with entry $K_{ij} = k\big(\bm{x}^{(i)},\bm{x}^{(j)}\big)$ for $i,j = 1,2,\cdots,n$. This conditional distribution is called the posterior probability distribution. The hyperparameters in the prior mean and covariance are often determined by the maximum  likelihood estimate or the maximum a posterior estimate~\cite{Frazier_2018}.

\begin{figure}
	\centering
	\includegraphics[width=0.6\linewidth]{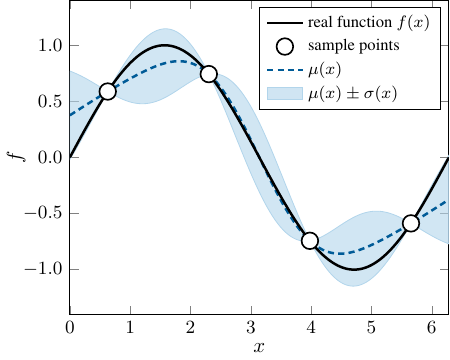}
	\caption{GP approximation of the $f = \sin(x)$ function. }
	\label{fig_GP}
\end{figure}

The GP model of a one-dimensional function $f = \sin(x)$ is illustrated in Fig.~\ref{fig_GP}. In this figure, the solid line represents the real function to be approximated, the circles represent the sample points, the dashed line represents the mean of the GP model and the filled area around the dashed line shows the standard derivation function of the GP model. We can see from the figure that the GP mean interpolates the samples. The standard derivation is zero at sample points and rises up between them.

\subsection{Acquisition Function}
After training the Gaussian process model, the next thing is to decide which point should be selected for expensive evaluation. The criterion for querying new samples is often called acquisition function or infill sampling function. Popular acquisition functions in Bayesian optimization are the expected improvement (EI)~\cite{Mockus_1975,Mockus_1994}, probability of improvement (PI)~\cite{Jones_2001}, lower confidence bound (LCB)~\cite{Jones_2001},  knowledge gradient (KG)~\cite{Frazier_2008,Frazier_2009}, entropy search (ES)~\cite{Hennig_2012}, predictive entropy search (PES)~\cite{Hernandez_2014} and so on. Among them,  EI is arguably the most widely used~\cite{Zhan_2020}.

\begin{figure}
	\centering
	\includegraphics[width=0.6\linewidth]{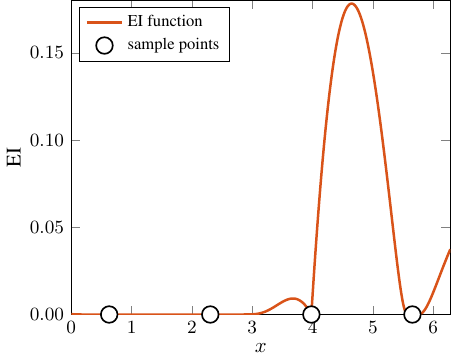}
	\caption{The EI function on the $f=\sin(x)$ problem.}
	\label{fig_EI}
\end{figure}

Assume current best solution  among the $n$ samples is $\bm{x}^{\star}$ and the corresponding minimum objective value is $f^{\star} = f(\bm{x}^{\star})$. The EI measures the expected value of improvement that the new point $\bm{x}$ can get beyond the current best solution $\bm{x}^{\star}$
\begin{equation}
	\text{EI}(\bm{x}) = \mathbb{E}\big[\max \big(f^{\star} - f(\bm{x}),0\big)\big].
\end{equation}
The closed-form expression can be derived using integration by parts~\cite{Jones_1998}
\begin{equation}
	\text{EI}(\bm{x})  = \big( f^{\star} - \mu(\bm{x}) \big) \Phi \bigg(\frac{f^{\star}-\mu(\bm{x})}{\sigma(\bm{x})}\bigg) + \sigma(\bm{x}) \phi \bigg(\frac{f^{\star}-\mu(\bm{x})}{\sigma(\bm{x})}\bigg)
\end{equation}
where $\Phi$ and $\phi$ are the cumulative and density distribution function of the standard Gaussian distribution respectively, and $\mu(\bm{x})$ and $\sigma(\bm{x})$ are the GP mean and standard deviation in (\ref{GP_mean}) and (\ref{GP_variance}) respectively.

The corresponding EI function of the GP model of the one-dimensional function is shown in Fig.~\ref{fig_EI}. We can see that the EI function is multi-modal. The value is zero at sample points and rises up between different samples. In this simple example, the next point the EI function locates is very close to the global optimum point of the real function.

\subsection{Bayesian Optimization}

Often, Bayesian optimization (BO) starts with an initial design of experiment (DoE) to get a set of samples and evaluates these initial samples with the real objective function. Then, Bayesian optimization goes into the sequential optimization process. In each iteration, a GP model is trained using the samples in current data set. After that, a new point is located by optimizing an acquisition function, and then evaluated with the real objective function. This newly evaluated sample is then added to the data set. This iteration process stops when the maximum number of function evaluations is reached. The computational framework of Bayesian optimization can be summarized in Algorithm~\ref{algorithm_BO}, where the EI acquisition function is used for example.

\begin{algorithm}
	\caption{Computational Framework of BO}
	\label{algorithm_BO}
	\begin{algorithmic}[1]	
		\REQUIRE $n_{\text{init}} = $ number of initial samples; 
		$n_{\max} = $ maximum number of  objective function evaluations. 
		\ENSURE best found solution $(\bm{x}^{\star},f^{\star})$.		
		\STATE \textbf{Design of experiment:} generate $n_{\text{init}}$ samples, evaluate them with the objective function, set current data set $\mathcal{D} = \{(\bm{x}^{(1)}, f^{(1)}),\cdots, (\bm{x}^{(n_{\text{init}})}, f^{(n_{\text{init}})})\}$, and set current number of evaluations $n = n_{\text{init}}$.
		\STATE \textbf{Best solution setup:} set current best solution as $\bm{x}^{\star} = {\underset {1 \le i \le n } {\arg\min}} f(\bm{x}^{(i)}) $ and $f^{\star} = {\underset {1 \le i \le n } {\min}} f(\bm{x}^{(i)})$.
		\WHILE{$n < n_{\max}$} 
		\STATE \textbf{GP training:} train a GP model using the current data set $\mathcal{D}$.
		\STATE \textbf{Infill selection:} locate a new point by maximizing the acquisition function
		\begin{equation*}
			\bm{x}^{(n+1)} = {\underset {\bm{x} \in \mathcal{X}} {\arg\max}} ~\text{EI}(\bm{x}).
		\end{equation*}    		
		\STATE \textbf{Expensive evaluation:} evaluate the new solution $f^{(n+1)} = f(\bm{x}^{(n+1)} )$, update data set $\mathcal{D} = \{\mathcal{D}, (\bm{x}^{(n+1)}, f^{(n+1)})\}$, and update current number of evaluations $n = n+1$. 
		\STATE  \textbf{Best solution update:} update current best solution as $\bm{x}^{\star} = {\underset {1 \le i \le n } {\arg\min}} f(\bm{x}^{(i)}) $ and $f^{\star} = {\underset {1 \le i \le n } {\min}} f(\bm{x}^{(i)})$.
		\ENDWHILE
	\end{algorithmic}	
\end{algorithm}

As can be seen, in each iteration the new point is selected by maximizing the EI function which has the same dimension as the objective function. When the dimension of the problem is lower than 20, the EI maximum can be efficiently located by using state-of-the-art evolutionary algorithms. However, when the dimension of the problem goes up to near 100, locating a good point will be a great challenge. In addition, the EI function often has highly multi-modal surface with lots of flat regions, which makes searching the high-dimensional EI maximum more difficult.

\section{Proposed Approach}
\label{section_coordinate_descent}
In this work, we propose the expected coordinate improvement based Bayesian optimization (ECI-BO) algorithm to tackle this problem. Instead of searching the high-dimensional space for identifying a new point, we locate a new point by searching one coordinate at a time. Through iterating over all the coordinates, the original problem can be solved gradually. The proposed algorithm is introduced in detail in the following.

\subsection{Expected Coordinate Improvement}

First, we propose the expected coordinate improvement (ECI) criterion to measure how much improvement we can get along one coordinate.  The origin of the proposed ECI is the traditional EI criterion, and the proposed ECI can be seen as an extension of the standard EI for measuring the improvement in coordinates. Assume $\bm{x}^{\star}$ is current best point and $f^{\star}$ is the corresponding current minimum objective. The current best solution is then a known $d$-dimensional vector
\begin{equation*}
\bm{x}^{\star}= [x_1^{\star},\cdots,x_{i-1}^{\star},x_{i}^{\star},x_{i+1}^{\star},\cdots,x_d^{\star}] 
\end{equation*}
where $d$ is the number of variables.  Then, the ECI along the $i$th coordinate is
\begin{equation}
	\label{eq_ECI}
	\text{ECI}_i(x) =  \big(f^{\star} -\mu(\bm{z}) \big) \Phi \bigg(\frac{f^{\star} - \mu(\bm{z})}{\sigma(\bm{z})}\bigg) + \sigma(\bm{z}) \phi \bigg(\frac{\mu(f^{\star} -\bm{z})}{\sigma(\bm{z})}\bigg)
\end{equation}
where 
\begin{equation*}
\bm{z}= [x_1^{\star},\cdots,x_{i-1}^{\star},x,x_{i+1}^{\star},\cdots,x_d^{\star}],  
\end{equation*}
$\Phi()$ and $\phi()$ are the cumulative and density distribution function of the standard Gaussian distribution respectively, and $\mu()$ and $\sigma()$ are the GP mean and standard deviation respectively. Comparing the vector $\bm{z}$ with $\bm{x}^{\star}$, we can find the the vector $\bm{z}$  has the same coordinates as the current best point $\bm{x}^{\star}$ except for the $i$th coordinate, which is the variable to be optimized in ECI. Therefore, the $	\text{ECI}_i(x)$ is the amount of expected improvement when we move the current best point $\bm{x}^{\star}$ along the $i$th coordinate.

\begin{figure}
	\centering
	\includegraphics[width=0.6\linewidth]{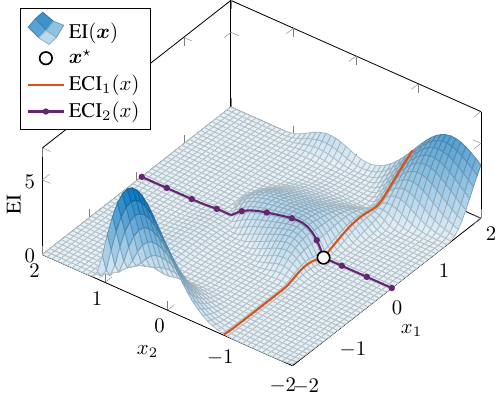}
	\caption{EI function and the corresponding ECI functions. }
	\label{fig_ECI}
\end{figure}

In fact, the ECI function is  a one-dimensional slice of the standard EI function while fixing all the other dimensions. A two-dimensional example is given in Fig.~\ref{fig_ECI}, where the surface is the standard EI function, the open circle is the current best point, and the two solid lines are the two ECI functions respectively. We can see that the $\text{ECI}_1(x)$ is the EI function fixing $x_2 = x_2^\star$, and the $\text{ECI}_2(x)$ is the EI function fixing $x_1 = x_1^\star$. As a result, the ECI function is able to measure the amount of improvement when we move the current best point along one coordinate.

The major advantage of the proposed ECI function over the traditional EI function is that it is always a one-dimensional function no matter how many dimensions the original problem has. As the dimension increases, the difficulty for optimizing the traditional EI increases exponentially while the difficulty for optimizing ECI criterion increases very slowly. Therefore, the ECI function has more advantages in solving high-dimensional optimization problems.

\subsection{Expected Coordinate Improvement based Bayesian Optimization}
Based on the ECI criterion, we propose the expected coordinate improvement based Bayesian optimization (ECI-BO) algorithm. The idea is to improve the current best point along one coordinate at a time, and iterates through all the coordinates as the optimization process goes on. To find the best order to optimize the coordinates, we use the maximal ECI values to approximately represent the effectiveness of the corresponding coordinates, and optimize the coordinates based on the order of their maximal ECI values. The computational framework of the proposed ECI-BO is given in Algorithm~\ref{algorithm_ECI_BO}.

\begin{algorithm}
	\caption{Computational Framework of the proposed ECI-BO}
	\label{algorithm_ECI_BO}
	\begin{algorithmic}[1]	
		\REQUIRE $n_{\text{init}} = $ number of initial samples; 
		$n_{\max} = $ maximum number of  objective function evaluations.
		\ENSURE best found solution $(\bm{x}^{\star},f^{\star})$.		
		\STATE \textbf{Design of experiment:} generate $n_{\text{init}}$ samples, evaluate them with the objective function, set current data set $\mathcal{D} = \{(\bm{x}^{(1)}, f^{(1)}),\cdots, (\bm{x}^{(n_{\text{init}})}, f^{(n_{\text{init}})})\}$, and set current number of evaluations $n = n_{\text{init}}$. 
		\STATE  \textbf{Best solution setup:} set current best solution as $\bm{x}^{\star} = {\underset {1 \le i \le n } {\arg\min}} f(\bm{x}^{(i)}) $ and $f^{\star} = {\underset {1 \le i \le n } {\min}} f(\bm{x}^{(i)})$.
		\WHILE{$n < n_{\max}$} 
		\FOR{$i = 1$ to $d$}
		\STATE \textbf{Calculating maximal ECIs:} Maximize the ECI function of the $i$th coordinate
		\begin{equation*}
			\text{ECI}_{i,\max} = \max \text{ECI}_i(x)
		\end{equation*}
		\ENDFOR			
		\STATE \textbf{Sorting maximal ECIs:} Sort the maximal ECI values in descending order to get the sorting indices
		\begin{equation*}
			[r_1,r_2,\cdots,r_d] = {\underset {1 \le i \le d} {\arg \text{sort}}}~\text{ECI}_{i,\max}  
		\end{equation*}

		\FOR{$i = 1$ to $d$}
		\STATE \textbf{GP training:} train a GP model using the current data set $\mathcal{D}$.
		
		\STATE \textbf{Infill selection:} maximize the ECI function along the $r_i$ coordinate
		\begin{equation*}
			x^{(n+1)} = {\underset {a_{r_i} \le x \le b_{r_i}} {\arg\max}} ~\text{ECI}_{r_i}(x)
		\end{equation*}
		and form the new solution as 
		\begin{equation*}
			\bm{x}^{(n+1)}= [x_1^{\star},\cdots, x_{r_i-1}^{\star},x^{(n+1)},x_{r_i+1}^{\star},\cdots,x_d^{\star}].
		\end{equation*}
		\STATE \textbf{Expensive evaluation:}  evaluate the new solution $f^{(n+1)} = f(\bm{x}^{(n+1)})$, update data set $\mathcal{D} = \{\mathcal{D}, (\bm{x}^{(n+1)}, f^{(n+1)})\}$, and update current number of evaluations $n = n+1$.
		\STATE \textbf{Best solution update:} update current best solution as $\bm{x}^{\star} = {\underset {1 \le i \le n } {\arg\min}} f(\bm{x}^{(i)}) $ and $f^{\star} = {\underset {1 \le i \le n } {\min}} f(\bm{x}^{(i)})$.
		\ENDFOR
		\ENDWHILE
	\end{algorithmic}	
\end{algorithm}

The key steps of the proposed ECI-BO algorithm are described in the following.
\begin{enumerate}
	\item \emph{Design of experiment:} in the first step, we generate $n_{\text{init}}$ initial samples using a design of experiment (DoE) method, such as Latin hypercube sampling method. After that, we evaluate these initial samples with the expensive objective function. In this step, these initial samples can be evaluated in parallel.
	\item \emph{Best solution setup:} the best solution of the  initial samples is identified as $(\bm{x}^{\star},f^{\star})$.
	\item \emph{Calculating maximal ECIs:} before the acquisition function optimization process, we try to find the best order to optimize the coordinates. First, we maximize the ECI function of all the $d$ coordinates. Note that these $d$ optimization problems are all one-dimensional and can be executed in parallel. 
	\item \emph{Sorting maximal ECIs:} After obtaining the maximal ECI values of all the coordinates, we sort the coordinates from the highest maximal ECI value to the lowest maximal ECI value, and  set the coordinate optimization order as the sorting order.  For example, if the maximal ECI values of a five-dimensional problems are $[200,300,500,400,100]$, then the coordinate optimization order is $\bm{r} = [3,4,2,1,5]$. 
	\item \emph{GP training:} all the samples in current data set are used for training the GP model. We train the GP model in the original space instead of the one-dimensional subspace to ensure that the model has reasonable accuracy for the prediction.
	\item \emph{Infill selection:}  in the $i$th iteration of the coordinate optimization process, we optimize the ECI function of the $r_i$th coordinate, where $r_i$ is the $i$th element of the previous obtained coordinate optimization order. Then, the infill solution is formed by replacing the $r_i$th coordinate of current best solution by the optimized value.
	\item \emph{Expensive evaluation:}  the new infill solution is evaluated with the expensive objective function and this newly evaluated point is then added to the data set.
	\item \emph{Best solution update:} finally we update the best solution by considering the newly evaluated point. At this point, one coordinate optimization process is finished. Then, we move to the next coordinate and repeat the process from Step 9 to Step 12. Once all the coordinates  has been covered, we go back to Step 4 to Step 7 to generate a new coordinate optimization order. This process continues until the maximum number of function evaluations is reached. 	
\end{enumerate}

From the above process, we can find that the major difference between the proposed ECI-BO and the standard BO is that the proposed approach maximizes the one-dimensional ECI function to query a new point for expensive evaluation instead of the $d$-dimensional EI function. The one-dimensional ECI function is often significantly easier to solve than the $d$-dimensional EI function. Through iterating through all the coordinates, the original problem can be gradually solved.

\subsection{A Two-Dimensional Example}

\begin{figure*}
	\centering
	\subfloat[real function and the GP approximation]{\includegraphics[width=0.48\linewidth]{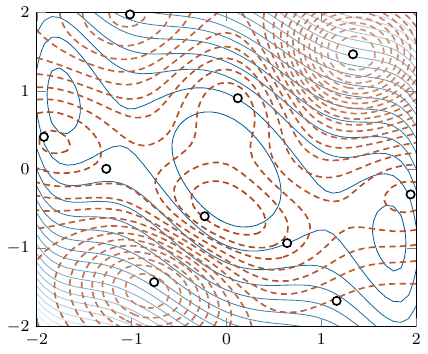}} 	\hfil
	\subfloat[the EI function and the ECI function]{\includegraphics[width=0.48\linewidth]{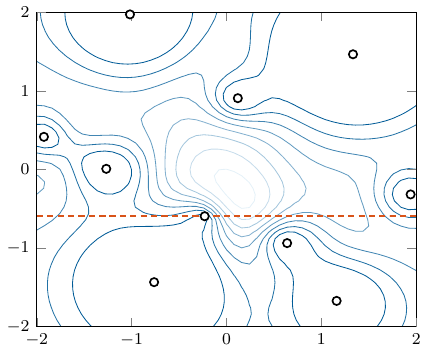}} \\
	\subfloat[search trajectory of BO]{\includegraphics[width=0.48\linewidth]{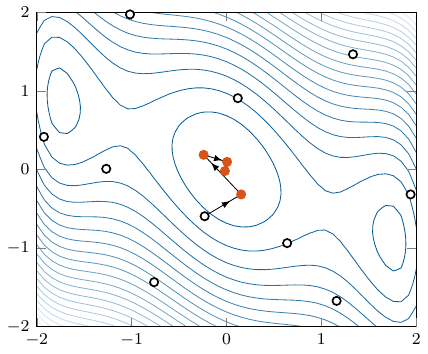}} 	\hfil
	\subfloat[search trajectory of ECI-BO]{\includegraphics[width=0.48\linewidth]{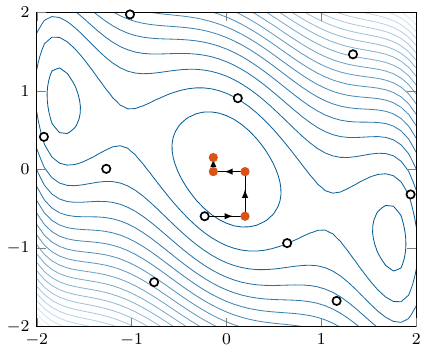}} \\
	\caption{Search trajectories of the standard BO and the proposed ECI-BO on the three-hump camel function.}
	\label{fig_search}
\end{figure*}

Since we move the current best solution along one coordinate at a time, the search trajectory of the proposed ECI-BO is very different from that of the standard BO. We demonstrate the search trajectories of the standard BO and the proposed ECI-BO  on the two-dimensional three-hump camel  function
\begin{equation}
	f = 2x_1^2 - 1.05x_1^4 + \frac{x_1^6}{6} + x_1x_2 + x_2^2
\end{equation}
where $-2 \le x_1 \le 2$ and $-2 \le x_2 \le 2$. Clearly, the two variables $x_1$ and $x_2$ are correlated in this example.

In Fig.~\ref{fig_search}(a), the solid contour is the real objective function, the open cycles are the ten initial samples, and the dashed contour is the GP approximation trained by the ten initial samples. We can see that the GP model can capture some trends of the real function, but there are still some differences between them. In Fig.~\ref{fig_search}(b), we show the current EI function using the solid contour and the ECI function long the $x_1$ coordinate using the dashed line. Fig.~\ref{fig_search}(c) and  Fig.~\ref{fig_search}(d) show the search trajectories of the standard BO approach and the proposed ECI-BO approach, respectively. We can clearly see the difference in the search pattern between them. In each iteration, the standard BO searches in the two-dimensional space to find the a new query point. In comparison, our proposed ECI-BO searches along one coordinate to locate a new point in one iteration. At the end of four iterations, both algorithms find solutions that are very close to the global optimum.

\section{Numerical Experiments}
\label{section_experiment}
In this section, we conduct numerical experiments to study the performance of our proposed ECI-BO algorithm. First, we compare our ECI-BO with the standard BO and five high-dimensional BOs to see whether the proposed approach has improvement over current BO approaches. Then, we compare the proposed ECI-BO with six surrogate-assisted evolutionary algorithms (SAEAs) to see how well our algorithm performs when compared with the state-of-the-art SAEAs. The Matlab implementation of our ECI-BO used in the experiments is available at  \url{https://github.com/zhandawei/Expected_Coordinate_Improvement}.

\subsection{Experiment Settings}
We use the Ellipsoid, Rosenbrock, Ackley, Griewank, Rastrigin problems to test the compared algorithms. The dimensions of these analytical problems are set to $d=30, 50, 100$ and $200$, respectively. In addition, we include the CEC 2013 test suite~\cite{Liang_2013} and the CEC 2017 test suit~\cite{Awad_2017} as the benchmark problems. The CEC 2013 test suit have 28 problem, in which $f_1$ to $f_5$ are unimodal problems, $f_6$ to $f_{20}$ are multimodal problems, and $f_{21}$ to $f_{28}$ are composition problems. The CEC 2017 test suit contains 29 problems, in which $f_1$ and $f_3$ are unimodal problems, $f_4$ to $f_{10}$ are simple multimodal problems, $f_{11}$ to $f_{20}$ are hybrid problems, and $f_{21}$ to $f_{30}$ are composition problems.  The dimensions of both CEC 2013 and CEC 2017 problems are set to $d=30, 50$ and $100$.  The maximal number of function evaluations is set to $n_{\max} = 1000$ for all test problems. 

We use the squared exponential (SE) function as the kernel function for the GP model. The optimal hyperparameter of the SE kernel is found within $[0.01,100]$ using the sequential quadratic programming algorithm. A real-coded genetic algorithm (GA) is applied to optimize the acquisition functions in the infill selection process. We use tournament selection for the parent selection, simulated binary crossover for the crossover and the polynomial mutation for the mutation respectively in the GA. Both the distribution indices for crossover and mutation are set to 20. The number of acquisition function evaluations is set to $200d$ throughout the experiments. For the standard BO, the population size of the GA is set to $2d$ and the number of maximal generation  is set to 100. For the proposed ECI-BO, the population size and the number of maximal generation are set to 10 and 20 respectively to maximize the 1-D ECI function.

All experiments are run 30 times with 30 different sets of initial samples.	The initial design points are the same for different algorithms in one run but are different for different runs. All experiments are conducted on a Window 10 machine with an Intel i9-10900X CPU and 64 GB RAM.

\begin{figure*}	
\centering
\makebox[\textwidth]{\parbox{1.2\textwidth}{
			\subfloat[30-D Ellipsoid]{\includegraphics[width=0.24\linewidth]{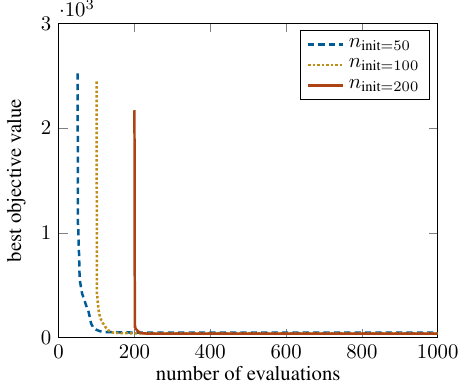}} 	\hfil
			\subfloat[50-D Ellipsoid]{\includegraphics[width=0.24\linewidth]{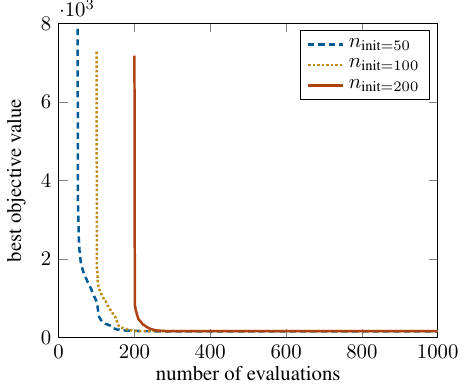}}\hfil
			\subfloat[100-D Ellipsoid]{\includegraphics[width=0.24\linewidth]{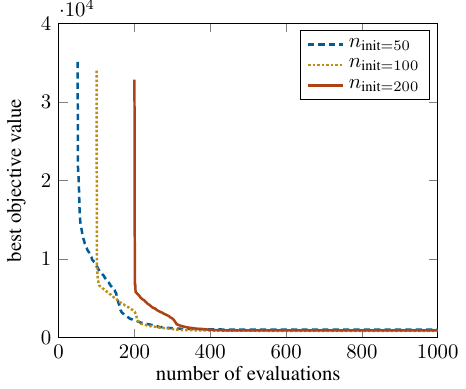}}\hfil
			\subfloat[200-D Ellipsoid]{\includegraphics[width=0.24\linewidth]{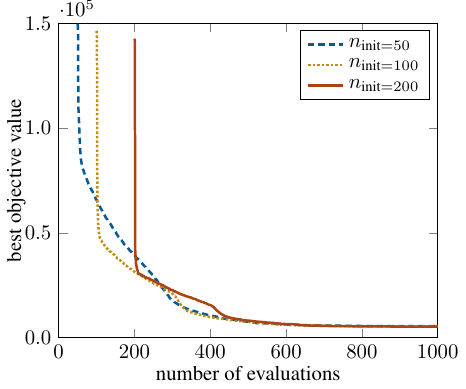}}  \vspace{-3mm}  \\ 
			\subfloat[30-D Rosenbrock]{\includegraphics[width=0.24\linewidth]{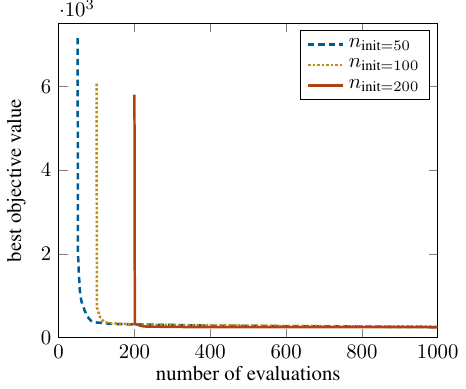}} 	\hfil
			\subfloat[50-D Rosenbrock]{\includegraphics[width=0.24\linewidth]{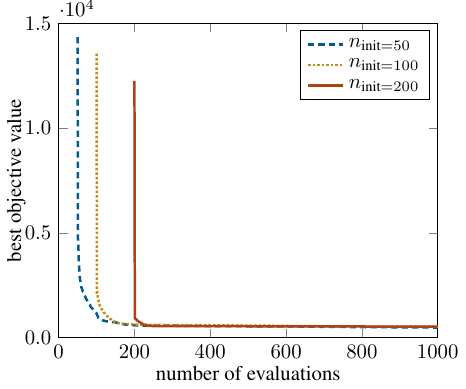}}\hfil
			\subfloat[100-D Rosenbrock]{\includegraphics[width=0.24\linewidth]{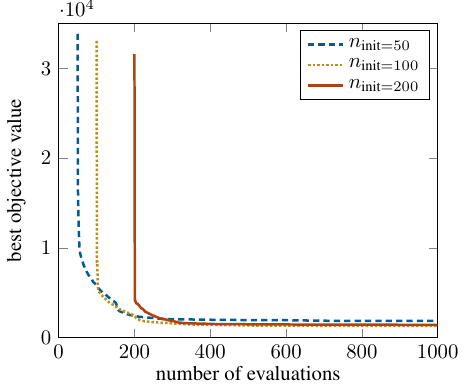}}\hfil
			\subfloat[200-D Rosenbrock]{\includegraphics[width=0.24\linewidth]{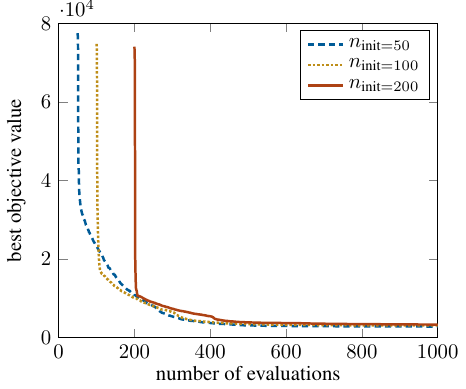}}  \vspace{-3mm}  \\
			\subfloat[30-D Ackley]{\includegraphics[width=0.24\linewidth]{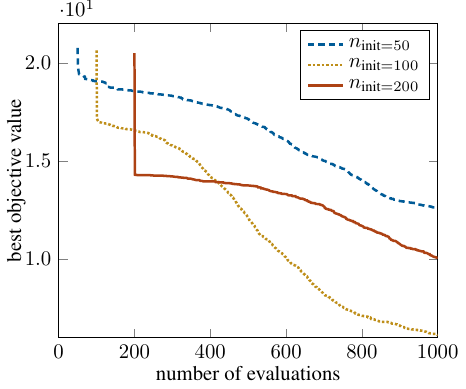}} 	\hfil
			\subfloat[50-D Ackley]{\includegraphics[width=0.24\linewidth]{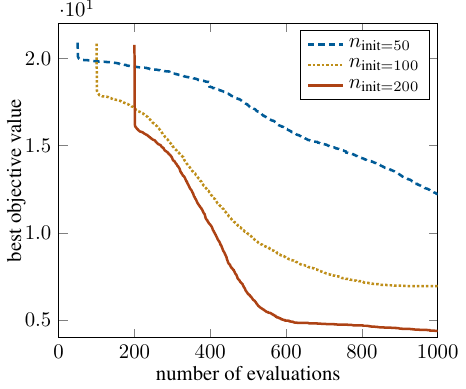}}\hfil
			\subfloat[100-D Ackley]{\includegraphics[width=0.24\linewidth]{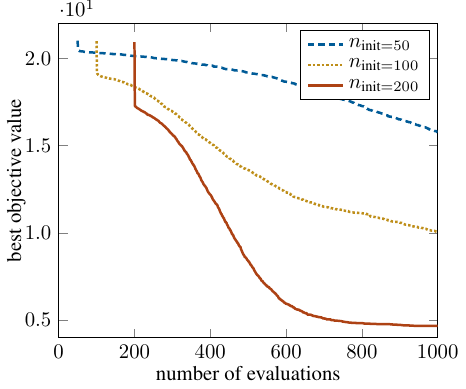}}\hfil
			\subfloat[200-D Ackley]{\includegraphics[width=0.24\linewidth]{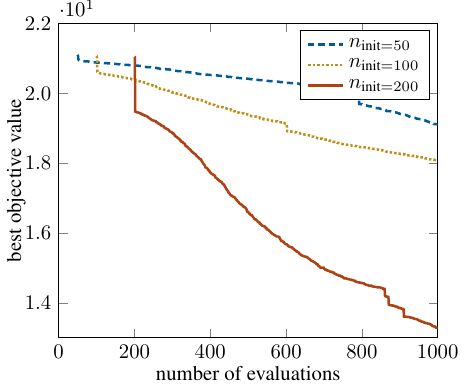}} \vspace{-3mm}  \\ 
			\subfloat[30-D Griewank]{\includegraphics[width=0.24\linewidth]{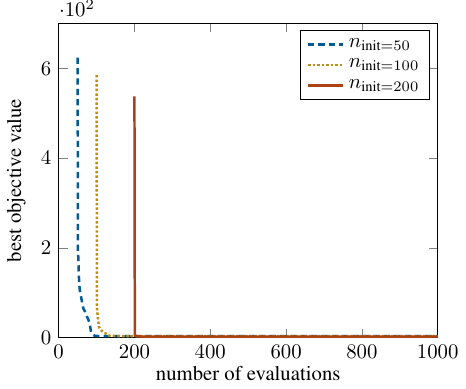}} 	\hfil
			\subfloat[50-D Griewank]{\includegraphics[width=0.24\linewidth]{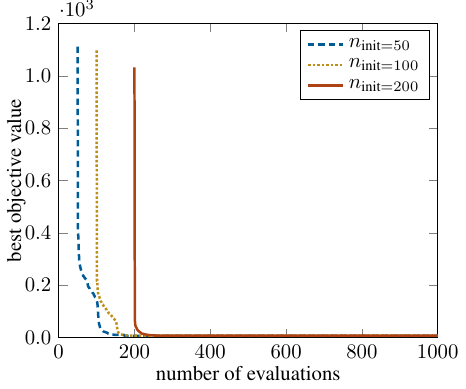}}\hfil
			\subfloat[100-D Griewank]{\includegraphics[width=0.24\linewidth]{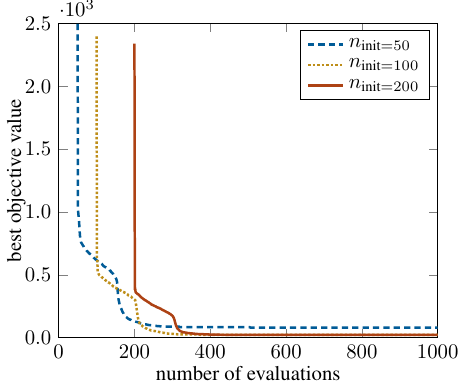}}\hfil
			\subfloat[200-D Griewank]{\includegraphics[width=0.24\linewidth]{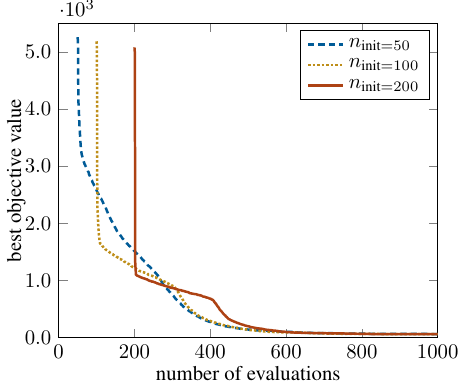}} \vspace{-3mm} \\ 
			\subfloat[30-D Rastrigin]{\includegraphics[width=0.24\linewidth]{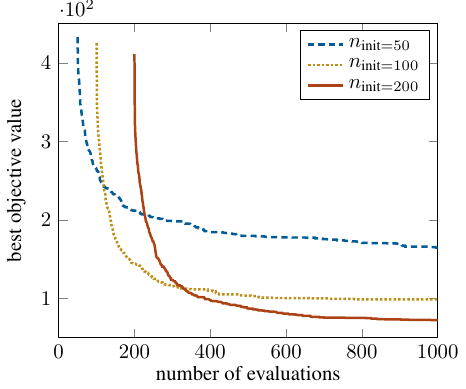}} 	\hfil
			\subfloat[50-D Rastrigin]{\includegraphics[width=0.24\linewidth]{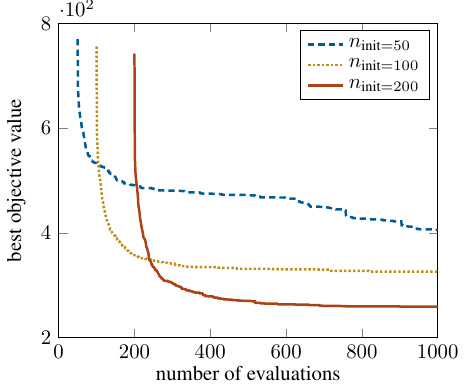}}\hfil
			\subfloat[100-D Rastrigin]{\includegraphics[width=0.24\linewidth]{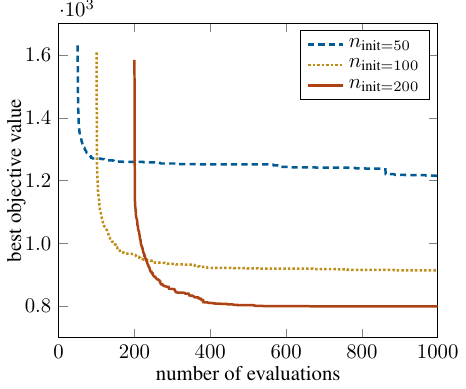}}\hfil
			\subfloat[200-D Rastrigin]{\includegraphics[width=0.24\linewidth]{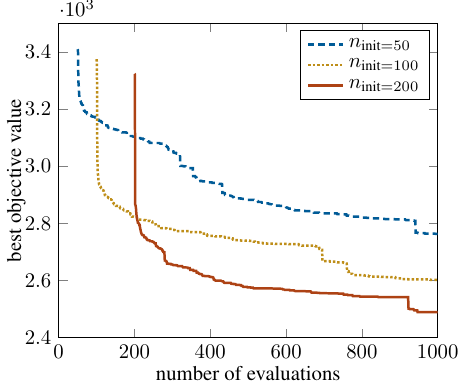}} \vspace{-3mm} \\ 
		}}
	\caption{Convergence plots of the standard BO with different number of initial samples.}
	\label{fig_number_initial}
\end{figure*}

The Latin hypercube sampling (LHS) method is used for generating the initial samples.  The number of initial samples $n_\text{init}$ is a hyperparameter for the BO algorithms.  When $n_\text{init}$ is small, the accuracy of the initial model is poor but we have more sequential samples in the iteration stage. In contrast, when $n_\text{init}$ is large, the accuracy of the initial model will be higher but the number of sequential samples will be lower. The hyperparameter $n_\text{init}$ takes the balance between these two factors. We conduct a set of experiments to select the best $n_\text{init}$ value. In the experiments, we run the standard BO algorithms on the Ellipsoid, Rosenbrock, Ackley, Griewank and Rastrigin problems by setting the  $n_\text{init}$  to 50, 100 and 200, respectively.  The convergence histories are plotted in Fig.~\ref{fig_number_initial}, where the average values of 30 runs are plotted. We can find that the number of initial samples has little effect on the Ellipsoid, Rosenbrock and Griewank problems, but has great effect on the Ackley and Rastrigin problems in terms of the final optimization results. On the 30-D Ackley problem, $n_\text{init}=100$ yields the best optimization results. However, on 50-D and higher-dimensional Ackley problems, BO with $n_\text{init}=200$ finds the best solutions. BO with $n_\text{init}=200$ also  performs  best on all the Rastrigin problems. It seems that it is helpful to use large number of initial samples to yield a more accurate GP model for the BO algorithm, especically on highly multi-modal problems like Ackley and Rastrigin. Based on the experiments results, we choose to set the hyperparameter $n_\text{init}$ to 200 in the following experiments.  

\subsection{Comparison with Bayesian Optimization Algorithms}
In the first set of experiments, we compare the proposed ECI-BO with six BOs, which are the standard BO, the Add-GP-UCB~\cite{Kandasamy_2015}, the Dropout~\cite{Li_2017}, the CoordinateLineBO~\cite{Kirschner_2019}, the TuRBO~\cite{Eriksson_2019}, and the MCTS-VS ~\cite{Song_2022}. The standard BO optimizes the acquisition function in the original design space to get new points for function evaluation. The Add-GP-UCB algorithm~\cite{Kandasamy_2015} decomposes the original problem into multiple low-dimensional sub-problems, solves these sub-problems and combines the solutions of the sub-problems into a high-dimensional solution. The low-dimensional space is also set to five in the experiments.  The open source code of Add-GP-UCB~\footnote{\url{https://github.com/kirthevasank/add-gp-banditss}} is used in the comparison. The Dropout approach~\cite{Li_2017} randomly selects a subset of variables to perform BO, and combine the solutions of the sub-problem with existing solutions to form a high-dimensional solution. Three different strategies for forming the high-dimensional solution are proposed in the work~\cite{Li_2017}, and the copy strategy is used in the comparison. The dimension of the subspace is also set to five in the experiments. The open source code  of the Dropout approach~\footnote{\url{https://github.com/licheng0794/highdimBO_dropout}} is used in the comparison. The CoordinateLineBO approach~\cite{Kirschner_2019} randomly selects a coordinate to optimize in each iteration. The coordinates of the infill solution are the same as the current best solution except for the optimized coordinate. We implement the CoordinateLineBO algorithm in Matlab according to the work~\cite{Kirschner_2019}. We use the genetic algorithm with population size of 10 and maximal generation of 20 to optimize the selected coordinate in each iteration for the CoordinateLineBO approach. The TuRBO~\cite{Eriksson_2019} uses the trust region approach to improve local search ability of Bayesian optimization algorithm on high-dimensional problems. The Thompson sampling is employed in TuRBO to select a batch of candidates to evaluate in each iteration. The number of trust regions is set to five and the number of batch evaluations is set to ten in the experiment. The open source code of TuRBO~\footnote{\url{https://github.com/uber-research/TuRBO}} is used in the comparison. Finally, the MCTS-VS approach~\cite{Song_2022}  employees the Monte Carlo tree search method to iteratively to select a subset of variables to optimize in each iteration. The open source code of MCTS-VS~\footnote{\url{https://github.com/lamda-bbo/MCTS-VS}} is used in the comparison.

The  optimization results of the compared algorithms on the selected test problems are given in Tables~\ref{table_BO_analytical} to ~\ref{table_BO_CEC2017_100D}, where the average values of 30 runs are shown.  We conduct the Wilcoxon signed rank test with significance level of $\alpha=0.05$ to see whether the optimization results of the compared algorithms have significant difference. We use $+$, $-$ and $\approx$ to donate that the results of the proposed ECI-BO are significantly better than, worse than and similar to the results of the compared algorithms, respectively.

\begin{table*}
	\renewcommand{\arraystretch}{1.0}
	\caption{Average results obtained by the compared BO algorithms on the analytical test problems}
	\label{table_BO_analytical}
	\centering
	\resizebox{\textwidth}{!}{
\begin{tabular}{c c c c c c c  c c } 
		\toprule
		$f$   & $d$& BO &  Add-GP-UCB & Dropout  &  \makecell[c]{Coordinate\\LineBO} &  TuRBO & MCTS-VS &  ECI-BO   \\
		\midrule
			 \multirow{4}{*}{Ellipsoid}  &30& 3.85E+01 $+$ & 8.84E+01 $+$  &5.42E+01 $+$ &2.90E-04 $+$ &6.21E+00 $+$ &5.58E+02 $+$ &9.75E-05 \\ 
			 							&50 & 1.67E+02 $+$ & 3.05E+03 $+$ &1.96E+02 $+$ &6.93E-01 $+$& 2.64E+02 $+$& 2.04E+03 $+$ &2.40E-02 \\ 
			 						    &100& 9.08E+02 $+$ & 2.15E+04 $+$ &3.03E+03 $+$ &2.22E+02 $+$& 5.67E+03 $+$ &1.05E+04 $+$& 1.89E+01 \\ 
			 							&200& 5.22E+03 $+$ & 1.10E+05 $+$ &4.10E+04 $+$ &1.03E+04 $+$ &3.29E+04 $+$ &5.53E+04 $+$ &1.40E+03 \\ 
			\multirow{4}{*}{Rosenbrock}  &30& 2.46E+02 $+$ & 2.82E+03 $+$ &1.43E+02 $+$ &7.79E+01 $+$ &9.26E+01 $+$ &1.10E+03 $+$ &4.71E+01 \\ 
			 							&50& 5.22E+02 $+$ & 7.23E+03 $+$ &3.43E+02 $+$ &2.18E+02 $+$ &4.84E+02 $+$ &2.44E+03 $+$ &1.22E+02 \\ 
			 							&100& 1.43E+03 $+$ & 2.08E+04 $+$ &2.17E+03 $+$ &8.49E+02 $+$ &3.10E+03 $+$ &7.79E+03 $+$ &4.35E+02 \\
\bottomrule
	\end{tabular}
}
\end{table*}

\begin{table*}
	\renewcommand{\arraystretch}{1.0}
	\caption{Average results obtained by the compared BO algorithms on 30-D CEC 2013 test problems}
	\label{table_BO_CEC2013_30D}
	\centering
	\resizebox{\textwidth}{!}{
	\begin{tabular}{c c c c c c c  c c } 
		\toprule
		$f$   & $d$& BO &  Add-GP-UCB & Dropout  &  \makecell[c]{Coordinate\\LineBO} &  TuRBO & MCTS-VS &  ECI-BO   \\
		\midrule
$f_{1}$ & 30 & -1.02E+03 $+$ & 4.78E+04 $+$ & 2.24E+03 $+$ & -1.40E+03 $+$ & -1.23E+03 $+$ & 1.23E+04 $+$ & -1.40E+03 \\
$f_{2}$ & 30 & 1.38E+08 $+$ & 9.89E+08 $+$ & 1.29E+08 $+$ & 6.18E+07 $+$ & 2.15E+08 $+$ & 2.73E+08 $+$ & 3.42E+07 \\
$f_{3}$ & 30 & 1.32E+18 $+$ & 1.98E+18 $+$ & -8.34E+26 $-$ & 1.48E+11 $\approx$ & 1.08E+11 $-$ & 1.46E+11 $\approx$ & 6.15E+11 \\
$f_{4}$ & 30 & 1.38E+05 $\approx$ & 1.42E+05 $\approx$ & 1.35E+05 $-$ & 1.55E+05 $\approx$ & 1.27E+05 $-$ & 1.50E+05 $\approx$ & 1.54E+05 \\
$f_{5}$ & 30 & 8.18E+03 $+$ & 2.46E+04 $+$ & -1.94E+02 $+$ & 6.98E+01 $+$ & 3.37E+03 $+$ & 9.26E+03 $+$ & -5.37E+02 \\
$f_{6}$ & 30 & 3.73E+02 $+$ & 7.73E+03 $+$ & -3.31E+02 $+$ & -7.55E+02 $+$ & -2.11E+02 $+$ & 2.64E+02 $+$ & -8.06E+02 \\
$f_{7}$ & 30 & 6.23E+05 $+$ & 4.78E+05 $+$ & 2.12E+03 $+$ & -3.45E+02 $\approx$ & -4.54E+02 $+$ & -5.35E+02 $-$ & -4.78E+02 \\
\bottomrule
	\end{tabular}
}
\end{table*}

\begin{table*}
	\renewcommand{\arraystretch}{1.0}
	\caption{Average results obtained by the compared BO algorithms on 50-D CEC 2013 test problems}
	\label{table_BO_CEC2013_50D}
	\centering
	\resizebox{\textwidth}{!}{
		\begin{tabular}{c c c c c c c  c c } 
			\toprule
			$f$   & $d$& BO &  Add-GP-UCB & Dropout  &  \makecell[c]{Coordinate\\LineBO} &  TuRBO & MCTS-VS &  ECI-BO   \\
			\midrule
			$f_{1}$ & 50 & -4.00E+02 $+$ & 9.18E+04 $+$ & 9.41E+03 $+$ & -1.39E+03 $+$ & 2.58E+03 $+$ & 3.35E+04 $+$ & -1.40E+03 \\
			$f_{2}$ & 50 & 4.10E+08 $+$ & 2.53E+09 $+$ & 2.79E+08 $+$ & 1.75E+08 $+$ & 6.68E+08 $+$ & 7.34E+08 $+$ & 8.66E+07 \\
			$f_{3}$ & 50 & 7.11E+16 $+$ & 9.91E+16 $+$ & -1.90E+30 $-$ & 3.00E+12 $\approx$ & 3.67E+11 $-$ & 1.91E+11 $-$ & 1.28E+12 \\
			$f_{4}$ & 50 & 2.20E+05 $\approx$ & 2.37E+05 $\approx$ & 2.18E+05 $-$ & 2.41E+05 $\approx$ & 2.17E+05 $\approx$ & 2.37E+05 $\approx$ & 2.41E+05 \\
			$f_{5}$ & 50 & 9.32E+03 $+$ & 4.25E+04 $+$ & 1.43E+03 $-$ & 6.57E+02 $-$ & 9.57E+03 $+$ & 1.83E+04 $+$ & 2.48E+03 \\
			$f_{6}$ & 50 & 2.14E+01 $+$ & 1.22E+04 $+$ & 8.38E+01 $+$ & -5.92E+02 $+$ & 1.11E+03 $+$ & 1.18E+03 $+$ & -7.35E+02 \\
			$f_{7}$ & 50 & 1.73E+05 $+$ & 1.51E+05 $+$ & -5.61E+02 $\approx$ & 4.80E+00 $\approx$ & -4.39E+02 $-$ & -4.90E+02 $-$ & 3.21E+02 \\
\bottomrule
		\end{tabular}
	}
\end{table*}

\begin{table*}
	\renewcommand{\arraystretch}{1.0}
	\caption{Average results obtained by the compared BO algorithms on 100-D CEC 2013 test problems}
	\label{table_BO_CEC2013_100D}
	\centering
	\resizebox{\textwidth}{!}{
		\begin{tabular}{c c c c c c c  c c } 
			\toprule
			$f$   & $d$& BO &  Add-GP-UCB & Dropout  &  \makecell[c]{Coordinate\\LineBO} &  TuRBO & MCTS-VS &  ECI-BO   \\
			\midrule
			$f_{1}$ & 100 & 1.45E+03 $+$ & 2.69E+05 $+$ & 6.39E+04 $+$ & 1.92E+03 $+$ & 5.04E+04 $+$ & 9.64E+04 $+$ & -1.40E+03 \\
			$f_{2}$ & 100 & 1.50E+09 $+$ & 1.04E+10 $+$ & 1.06E+09 $+$ & 5.66E+08 $+$ & 2.60E+09 $+$ & 2.54E+09 $+$ & 3.48E+08 \\
			$f_{3}$ & 100 & 9.96E+23 $+$ & 2.81E+24 $+$ & -2.97E+33 $-$ & 8.26E+18 $\approx$ & 1.99E+19 $\approx$ & 5.32E+18 $\approx$ & 5.88E+19 \\
			$f_{4}$ & 100 & 5.09E+05 $\approx$ & 4.95E+05 $\approx$ & 4.78E+05 $-$ & 5.25E+05 $\approx$ & 4.21E+05 $-$ & 4.85E+05 $-$ & 5.23E+05 \\
			$f_{5}$ & 100 & 3.99E+04 $-$ & 1.22E+05 $+$ & 1.11E+04 $-$ & 2.67E+04 $-$ & 2.84E+04 $-$ & 4.01E+04 $\approx$ & 4.71E+04 \\
			$f_{6}$ & 100 & 2.68E+03 $+$ & 7.41E+04 $+$ & 9.30E+03 $+$ & 7.65E+02 $+$ & 8.53E+03 $+$ & 1.13E+04 $+$ & -9.13E+01 \\
			$f_{7}$ & 100 & 5.09E+08 $+$ & 1.08E+09 $+$ & 7.55E+05 $\approx$ & 9.98E+05 $\approx$ & 1.75E+06 $\approx$ & 9.56E+05 $\approx$ & 9.48E+05 \\
\bottomrule
		\end{tabular}
	}
\end{table*}

\begin{table*}
	\renewcommand{\arraystretch}{1.0}
	\caption{Average results obtained by the compared BO algorithms on 30-D CEC 2017 test problems}
	\label{table_BO_CEC2017_30D}
	\centering
	\resizebox{\textwidth}{!}{
		\begin{tabular}{c c c c c c c  c c } 
			\toprule
			$f$   & $d$& BO &  Add-GP-UCB & Dropout  &  \makecell[c]{Coordinate\\LineBO} &  TuRBO & MCTS-VS &  ECI-BO   \\
			\midrule
		$f_{1}$ & 30 & 1.21E+09 $+$ & 5.91E+10 $+$ & 4.77E+09 $+$ & 4.31E+04 $+$ & 5.00E+08 $+$ & 1.88E+10 $+$ & 1.06E+04 \\
		$f_{3}$ & 30 & 1.85E+05 $-$ & 2.33E+05 $-$ & 2.29E+05 $-$ & 2.76E+05 $\approx$ & 2.01E+05 $-$ & 2.49E+05 $\approx$ & 2.74E+05 \\
		$f_{4}$ & 30 & 2.83E+03 $+$ & 1.73E+04 $+$ & 1.38E+03 $+$ & 6.64E+02 $+$ & 2.01E+03 $+$ & 3.21E+03 $+$ & 5.80E+02 \\
		$f_{5}$ & 30 & 7.45E+02 $+$ & 9.82E+02 $+$ & 7.68E+02 $+$ & 7.07E+02 $+$ & 6.22E+02 $-$ & 7.54E+02 $+$ & 6.66E+02 \\
		$f_{6}$ & 30 & 6.34E+02 $+$ & 6.96E+02 $+$ & 6.31E+02 $+$ & 6.15E+02 $+$ & 6.14E+02 $+$ & 6.37E+02 $+$ & 6.06E+02 \\
		$f_{7}$ & 30 & 1.02E+03 $+$ & 2.31E+03 $+$ & 1.10E+03 $+$ & 9.84E+02 $+$ & 9.71E+02 $+$ & 1.40E+03 $+$ & 9.33E+02 \\
		$f_{8}$ & 30 & 1.03E+03 $+$ & 1.24E+03 $+$ & 1.06E+03 $+$ & 1.02E+03 $+$ & 9.16E+02 $-$ & 1.03E+03 $+$ & 9.45E+02 \\
\bottomrule
		\end{tabular}
	}
\end{table*}

\begin{table*}
	\renewcommand{\arraystretch}{1.0}
	\caption{Average results obtained by the compared BO algorithms on 50-D CEC 2017 test problems}
	\label{table_BO_CEC2017_50D}
	\centering
	\resizebox{\textwidth}{!}{
		\begin{tabular}{c c c c c c c  c c } 
			\toprule
			$f$   & $d$& BO &  Add-GP-UCB & Dropout  &  \makecell[c]{Coordinate\\LineBO} &  TuRBO & MCTS-VS &  ECI-BO   \\
			\midrule
		$f_{1}$ & 50 & 3.86E+09 $+$ & 1.49E+11 $+$ & 1.76E+10 $+$ & 5.32E+07 $+$ & 6.24E+09 $+$ & 4.84E+10 $+$ & 3.73E+05 \\
		$f_{3}$ & 50 & 4.28E+05 $-$ & 5.46E+05 $\approx$ & 4.32E+05 $-$ & 4.88E+05 $\approx$ & 3.54E+05 $-$ & 4.13E+05 $-$ & 4.89E+05 \\
		$f_{4}$ & 50 & 4.66E+03 $+$ & 4.58E+04 $+$ & 3.94E+03 $+$ & 1.21E+03 $+$ & 7.12E+03 $+$ & 7.42E+03 $+$ & 8.02E+02 \\
		$f_{5}$ & 50 & 9.05E+02 $+$ & 1.37E+03 $+$ & 1.06E+03 $+$ & 9.46E+02 $+$ & 8.85E+02 $+$ & 9.78E+02 $+$ & 7.88E+02 \\
		$f_{6}$ & 50 & 6.47E+02 $+$ & 7.11E+02 $+$ & 6.47E+02 $+$ & 6.26E+02 $+$ & 6.23E+02 $+$ & 6.50E+02 $+$ & 6.06E+02 \\
		$f_{7}$ & 50 & 1.32E+03 $+$ & 4.11E+03 $+$ & 1.55E+03 $+$ & 1.30E+03 $+$ & 1.34E+03 $+$ & 2.21E+03 $+$ & 1.15E+03 \\
		$f_{8}$ & 50 & 1.24E+03 $+$ & 1.68E+03 $+$ & 1.36E+03 $+$ & 1.24E+03 $+$ & 1.18E+03 $+$ & 1.28E+03 $+$ & 1.07E+03 \\
\bottomrule
		\end{tabular}
	}
\end{table*}

\begin{table*}
	\renewcommand{\arraystretch}{1.0}
	\caption{Average results obtained by the compared BO algorithms on 100-D CEC 2017 test problems}
	\label{table_BO_CEC2017_100D}
	\centering
	\resizebox{\textwidth}{!}{
		\begin{tabular}{c c c c c c c  c c } 
			\toprule
			$f$   & $d$& BO &  Add-GP-UCB & Dropout  &  \makecell[c]{Coordinate\\LineBO} &  TuRBO & MCTS-VS &  ECI-BO   \\
			\midrule
			$f_{1}$ & 100 & 1.16E+10 $+$ & 2.35E+11 $+$ & 9.60E+10 $+$ & 4.20E+09 $+$ & 8.32E+10 $+$ & 1.39E+11 $+$ & 2.76E+07 \\
			$f_{3}$ & 100 & 1.58E+06 $\approx$ & 4.45E+09 $\approx$ & 9.54E+05 $\approx$ & 9.63E+05 $\approx$ & 9.54E+05 $\approx$ & 2.36E+06 $+$ & 9.70E+05 \\
			$f_{4}$ & 100 & 1.44E+04 $+$ & 7.09E+04 $+$ & 1.67E+04 $+$ & 6.81E+03 $+$ & 3.24E+04 $+$ & 2.51E+04 $+$ & 3.37E+03 \\
			$f_{5}$ & 100 & 1.61E+03 $+$ & 2.21E+03 $+$ & 2.02E+03 $+$ & 1.69E+03 $+$ & 1.74E+03 $+$ & 1.67E+03 $+$ & 1.42E+03 \\
			$f_{6}$ & 100 & 6.63E+02 $+$ & 7.24E+02 $+$ & 6.72E+02 $+$ & 6.42E+02 $+$ & 6.55E+02 $+$ & 6.60E+02 $+$ & 6.12E+02 \\
			$f_{7}$ & 100 & 2.23E+03 $+$ & 4.49E+03 $+$ & 4.15E+03 $+$ & 2.35E+03 $+$ & 3.60E+03 $+$ & 4.73E+03 $+$ & 1.81E+03 \\
			$f_{8}$ & 100 & 1.93E+03 $+$ & 2.58E+03 $+$ & 2.37E+03 $+$ & 2.05E+03 $+$ & 2.05E+03 $+$ & 1.96E+03 $+$ & 1.71E+03 \\
\bottomrule
		\end{tabular}
	}
\end{table*}

First, we compare our proposed ECI-BO with the standard BO algorithm. From the tables, we can see that our proposed ECI-BO obtains significantly better results than the standard BO on most of the test problems.  Overall, our proposed ECI-BO finds better solutions on eighteen of the twenty analytical test problems, fifty-nine of the eighty-four CEC 2013 problems, and seventy-three of the eighty-seven CEC 2017 test problems . The major difference between these two algorithms is that the proposed ECI-BO searches along one coordinate to find an infill solution in each iteration while the standard BO searches in the original design space for finding an infill solution.  On high-dimensional problems, it is very challenging for the standard BO to find good solutions in a very large search space.  At the end of 1000 evaluations, our proposed ECI-BO is able to find better solutions on most of the test problems. The experiment results can empirically prove the effectiveness of the proposed expected coordinate improvement approach. 

 The Add-GP-UCB~\cite{Kandasamy_2015} assumes the objective function is a summation of multiple low-dimensional functions. However, most of the the selected problems do not meet this assumption. In comparison, our proposed ECI-BO does not make any structural assumption about the objective function we are optimizing. From the tables, we can find that the Add-GP-UCB can not achieve satisfying results on the test problems.  Our proposed ECI-BO performs significantly better than the Add-GP-UCB  on most of the benchmark problems. 

The Dropout approach~\cite{Li_2017} randomly selects a few variables to optimize in each iteration. When evaluating the new solutions, this approach uses the values of existing points to fill the unselected variables. The Dropout approach also trains the model in the selected subspace, which might result in poor GP models. In comparison, our approach trains the GP model in the original design space. This might be the reason why our proposed approach has better performance than the Dropout approach. From the tables we can find that our proposed ECI-BO performs significantly better on eighteen analytical problems, fifty-two CEC 2013 problems and sixty-five CEC 2017 problems compared with the Dropout approach. 

The CoordinateLineBO~\cite{Kirschner_2019} is the most similar approach to our proposed ECI-BO. Both the CoordinateLineBO and our ECI-BO select one variable to optimize in each iteration. However, there are two major differences between them. First, our proposed algorithm iterates the coordinates over permutations while the the CoordinateLineBO iterates the coordinates randomly. When we iterates $d$ times, all the variables will be covered by the proposed approach, but some variables may not be covered by the CoordindateLineBO approach since some variables might be selected multiple times.  Second, the CoordinateLineBO treats the variables equally, while our proposed ECI-BO learns the effectiveness of the variables based on their ECI values. The experiment results in the tables show that the proposed ECI-BO performs significantly better than the CoordinateLineBO approach on most of the test problems. This can empirically prove the effectiveness of the proposed expected coordinate improvement approach.

The TuRBO~\cite{Eriksson_2019} employees the trust region strategy in the BO framework to improve the local search ability of BO on high-dimensional problems. Although TuRBO reduces the search space to local regions, the local regions are still high-dimensional spaces. Searching in the high-dimensional local spaces is still challenging. Compared with the TuRBO, our ECI-BO finds better solutions on the majority of the test problems. This means our proposed expected coordinate improvement approach is more effective than the trust region approach on the test problems.

The MCTS-VS~\cite{Song_2022} is a recently developed variable selection strategy for high-dimensional optimization. It employees the Monte Carlo tree search method to partition the variables into important and unimportant ones, and only optimizes the important variables~\cite{Song_2022}.  For problems whose variables are all important, this approach might not be very efficient since most of the variables are not optimized after the variable selection. The experiment results show that our proposed ECI-BO performs significantly better than the MCTS-VS approach on most of the problems. This indicates that our ECI-BO is more suitable for problems whose variables are all effective to the objective function.

\begin{figure*}	
	\centering
	\makebox[\textwidth]{\parbox{1.2\textwidth}{
			\subfloat[30-D Ellipsoid]{\includegraphics[width=0.24\linewidth]{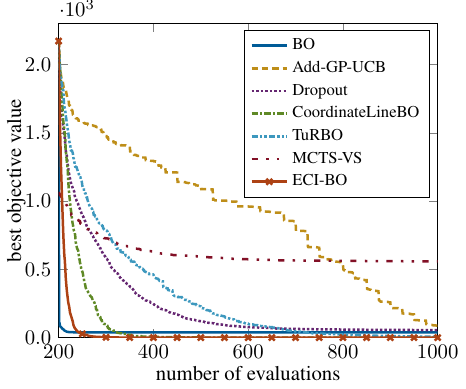}} 	\hfil
			\subfloat[50-D Ellipsoid]{\includegraphics[width=0.24\linewidth]{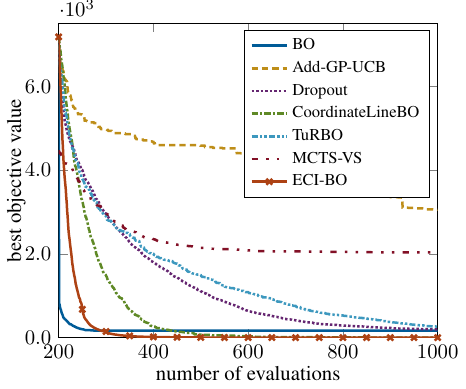}}\hfil
			\subfloat[100-D Ellipsoid]{\includegraphics[width=0.24\linewidth]{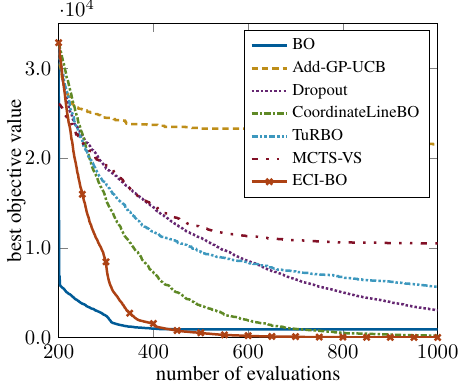}}\hfil
			\subfloat[200-D Ellipsoid]{\includegraphics[width=0.24\linewidth]{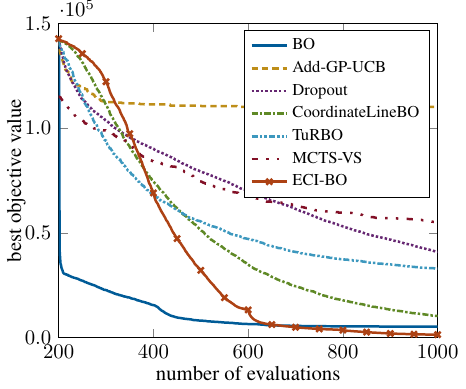}}   \vspace{-3mm} \\
			\subfloat[30-D Rosenbrock]{\includegraphics[width=0.24\linewidth]{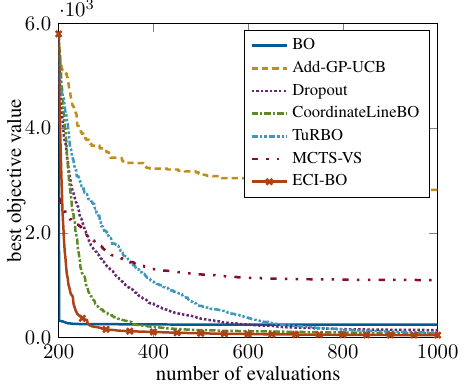}} 	\hfil
			\subfloat[50-D Rosenbrock]{\includegraphics[width=0.24\linewidth]{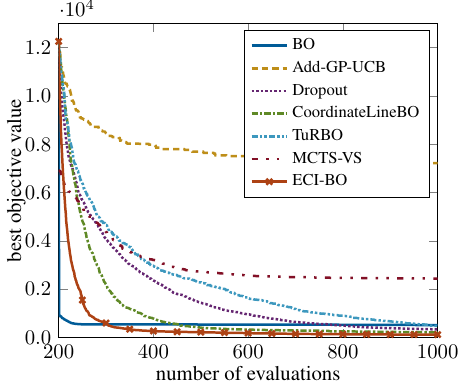}}\hfil
			\subfloat[100-D Rosenbrock]{\includegraphics[width=0.24\linewidth]{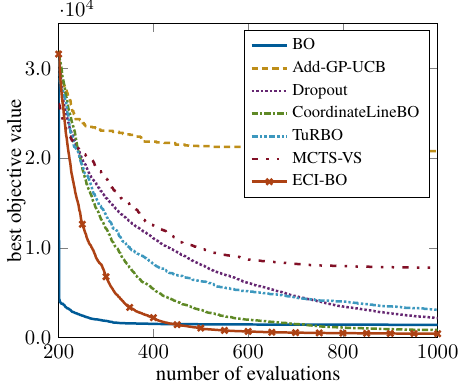}}\hfil
			\subfloat[200-D Rosenbrock]{\includegraphics[width=0.24\linewidth]{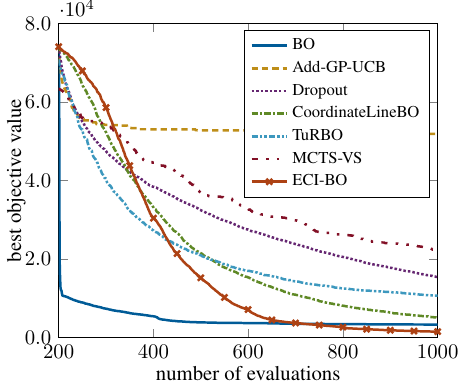}} \vspace{-3mm}  \\ 
			\subfloat[30-D Ackley]{\includegraphics[width=0.24\linewidth]{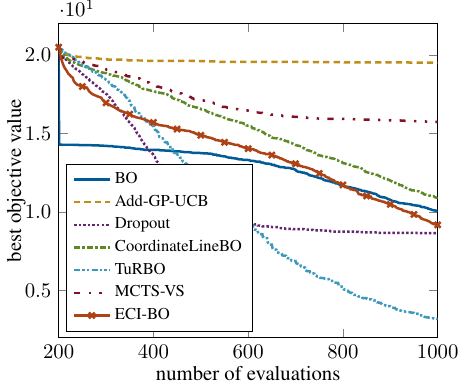}} 	\hfil
			\subfloat[50-D Ackley]{\includegraphics[width=0.24\linewidth]{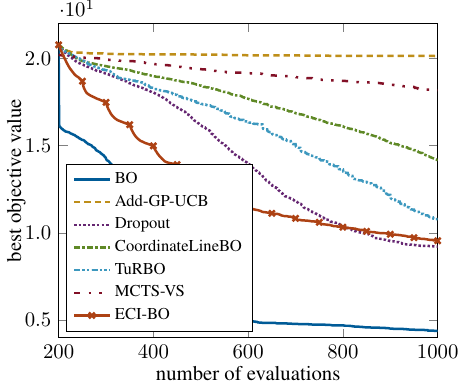}}\hfil
			\subfloat[100-D Ackley]{\includegraphics[width=0.24\linewidth]{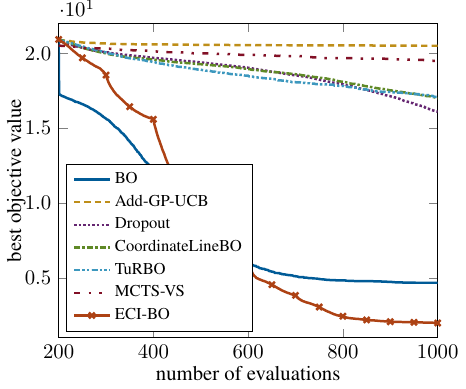}}\hfil
			\subfloat[200-D Ackley]{\includegraphics[width=0.24\linewidth]{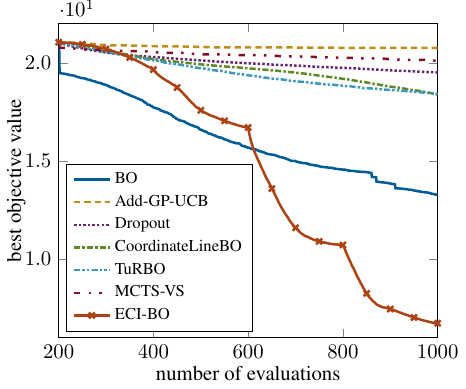}} \vspace{-3mm}  \\ 
			\subfloat[30-D Griewank]{\includegraphics[width=0.24\linewidth]{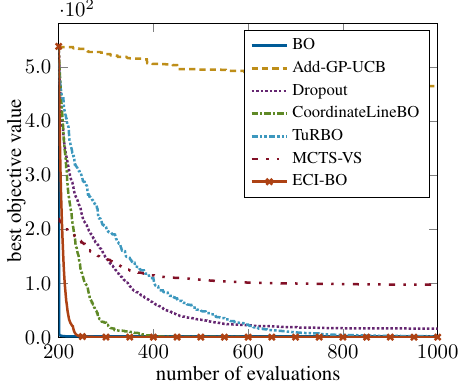}} 	\hfil
			\subfloat[50-D Griewank]{\includegraphics[width=0.24\linewidth]{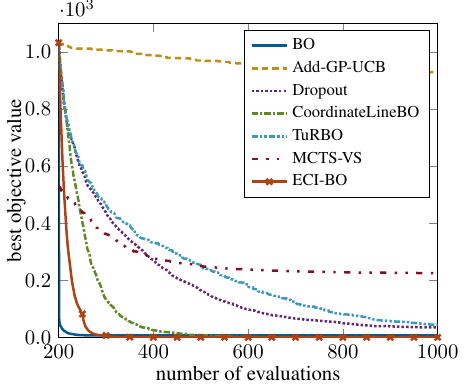}}\hfil
			\subfloat[100-D Griewank]{\includegraphics[width=0.24\linewidth]{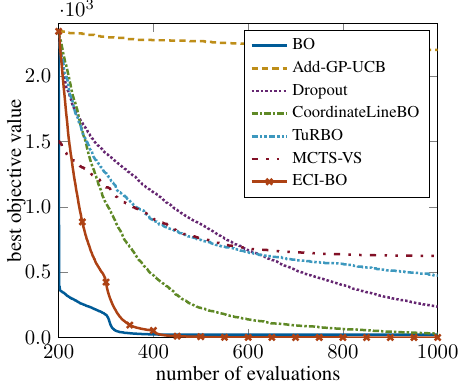}}\hfil
			\subfloat[200-D Griewank]{\includegraphics[width=0.24\linewidth]{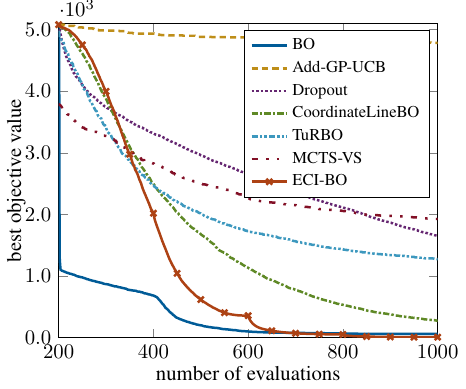}} \vspace{-3mm} \\ 
			\subfloat[30-D Rastrigin]{\includegraphics[width=0.24\linewidth]{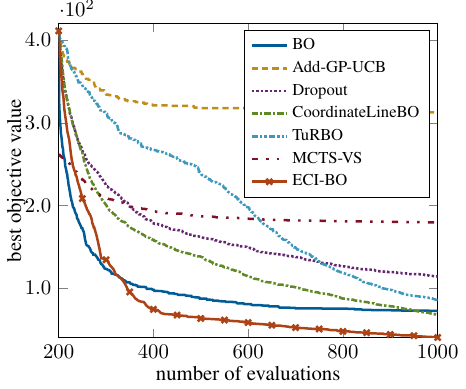}} 	\hfil
			\subfloat[50-D Rastrigin]{\includegraphics[width=0.24\linewidth]{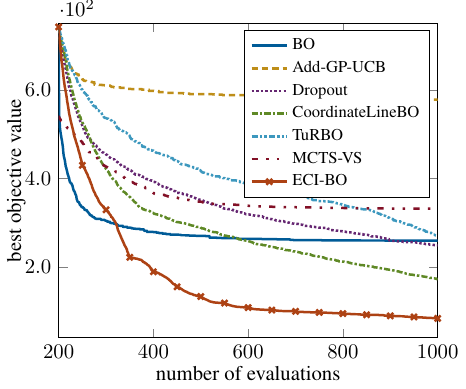}}\hfil
			\subfloat[100-D Rastrigin]{\includegraphics[width=0.24\linewidth]{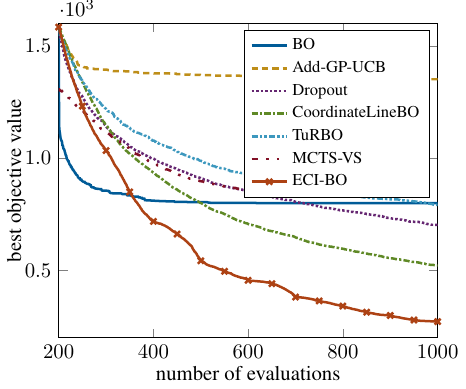}}\hfil
			\subfloat[200-D Rastrigin]{\includegraphics[width=0.24\linewidth]{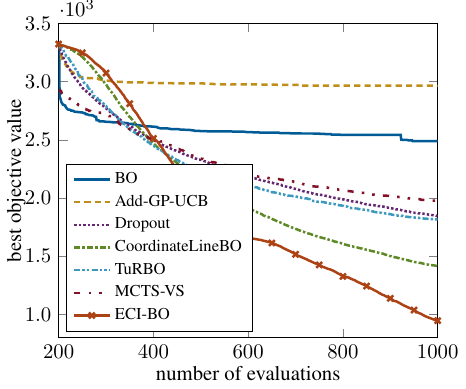}}  \vspace{-3mm}  \\
	}}
	\caption{Convergence plots of the compared BO algorithms on the analytical test problems.}
	\label{fig_comparison_BO}
\end{figure*}

The convergence histories of the compared algorithms on the twenty analytical test problems are plotted in Fig.~\ref{fig_comparison_BO}, where the average values of the 30 runs are shown. From the iteration histories we can find a very interesting phenomenon. The standard BO converges faster than the proposed ECI-BO at the beginning of iterations, but converges slower than the proposed ECI-BO at the end of the iterations on most of the test problems. The standard BO searches in the original space to find new solutions while the proposed ECI-BO searches along one coordinate. Therefore, the standard BO is able to find better solutions than the ECI-BO at the beginning of the iterations. However, as the iteration goes on, most of the promising areas have been explored and  exploitation is needed for further improvement. The standard BO still searches in the whole design space to find promising solutions, which often brings very little improvement as the iteration continues. On the contrary, the proposed ECI-BO searches along one coordinate at a time to find new solutions. Although it converges slower than the standard BO at the beginning of iterations when exploration is needed, it is able to continually find good solutions at the middle and at end of iterations when exploitation is urgently needed. At the end of iterations, the proposed ECI-BO often finds better solutions than the standard BO approach. Since the standard BO has faster convergence speed at the beginning of iterations, it is still preferred when the computation budget is extremely limited.  But when the computation budget is fairly enough, the proposed ECI-BO should be preferred. Compared with other high-dimensional BOs, we can see that the proposed ECI-BO converges fastest and achieves the best results at the end of 1000 evaluations on most of the problems.

\subsection{Comparison with Surrogate-Assisted Evolutionary Algorithms}
In this section, we compare the proposed ECI-BO with six surrogate-assisted evolutionary algorithms (SAEAs), which are also widely used for solving expensive optimization problems~\cite{Xie_2023,LiGH_2023,LiG_2023}. The compared SAEAs all use Gaussian process model as the surrogate model.  They are the Gaussian process-assisted differential evolution (GPDE) algorithm, the Gaussian process surrogate model assisted evolutionary algorithm for medium-scale computationally expensive optimization problems (GPEME)~\cite{Liu_2014}, the multiobjective infill criterion-driven Gaussian process-assisted social learning particle swarm optimization (MGP-SLPSO)~\cite{Tian_2019}, the incremental Kriging-assisted evolutionary algorithm (IKAEA)~\cite{Zhan_2021}, the Kriging-assisted evolutionary algorithm based on anisotropic expected improvement (KAEA-AEI)~\cite{Zhan_2023b}, and the linear subspace surrogate model-assisted evolutionary algorithm (L2SMEA)~\cite{Si_2023}. The GPDE uses the GP model as the surrogate and the DE algorithm as the base EA. In each generation of the GPDE approach, the individual with the highest EI value is selected for expensive evaluation. The GPEME~\cite{Liu_2014} also uses the GP model as the surrogate model, but it uses only the most recently 100 samples to train the GP model instead of using all the sample points to accelerate the training process. The lower confidence bound (LCB) criterion is utilized to select individuals for expensive evaluation.  The MGP-SLPSO~\cite{Tian_2019} utilizes the multiobjective criterion to select the individuals in social learning particle swarm optimization algorithm to do expensive evaluations.  The open source code of MGP-SLPSO~\footnote{\url{https://github.com/Jetina/MGPSLPSO}} is used in the experiments. The IKAEA~\cite{Zhan_2021} approach uses the incremental learning approach to training the GP model during the EA iterations. The open source code of IKAEA~\footnote{\url{https://github.com/zhandawei/Incremental_Kriging_Assisted_Evolutionary_Algorithm}} is employed in the comparison.  The KAEA-AEI~\cite{Zhan_2023b} approaches utilizes the anisotropic expected improvement criterion to select promising individuals for expensive evaluation. The open source code of KAEA-AEI~\footnote{\url{https://github.com/zhandawei/Anisotropy_Expected_Improvement}} is used in the experiments. The L2SMEA~\cite{Si_2023} trains multiple Kriging models on the selected linear subspaces, and uses the NSGA-III~\cite{Deb_2014} to optimizes all the subproblems simultaneously.  The open source code of L2SMEA in PlatEMO~\cite{Tian_2017}~\footnote{\url{https://github.com/BIMK/PlatEMO}} is used in the experiments.

\begin{table*}
	\renewcommand{\arraystretch}{1.0}
	\caption{Average results obtained by ECI-BO and the compared SAEAs on the analytical test porblems}
	\label{table_SAEA_analytical}
	\centering
		\resizebox{\textwidth}{!}{
\begin{tabular}{c c c c c c c  c c } 
		\toprule
		$f$   & $d$& GPDE &  GPEME & MGP-SLPSO  &  IKAEA &  KAEA-AEI & L2SMEA &  ECI-BO   \\
		\midrule
			\multirow{4}{*}{Ellipsoid}  & 30 & 2.20E+00 $+$ & 1.66E-01 $+$ & 8.20E-09 $-$ & 7.97E-03 $+$ & 6.58E-01 $+$ & 4.08E+02 $+$ & 9.75E-05 \\
			 & 50 & 6.41E+01 $+$ & 1.52E+01 $+$ & 2.59E-06 $-$ & 7.33E+00 $+$ & 3.21E+01 $+$ & 1.50E+03 $+$ & 2.40E-02 \\
			& 100 & 1.47E+03 $+$ & 8.42E+02 $+$ & 9.48E-01 $-$ & 6.40E+02 $+$ & 1.05E+03 $+$ & 8.27E+03 $+$ & 1.89E+01 \\
			& 200 & 1.50E+04 $+$ & 1.13E+04 $+$ & 3.78E+03 $+$ & 2.27E+04 $+$ & 1.20E+04 $+$ & 3.94E+04 $+$ & 1.40E+03 \\
			 \multirow{4}{*}{Rosenbrock}  & 30  & 8.58E+01 $+$ & 2.78E+01 $-$ & 4.93E+01 $\approx$ & 6.03E+01 $\approx$ & 6.84E+01 $+$ & 6.20E+02 $+$ & 4.71E+01 \\
			& 50  & 2.56E+02 $+$ & 1.07E+02 $\approx$ & 1.30E+02 $\approx$ & 2.18E+02 $+$ & 2.30E+02 $+$ & 1.52E+03 $+$ & 1.22E+02 \\
			& 100  & 9.57E+02 $+$ & 5.91E+02 $+$ & 4.94E+02 $+$ & 8.20E+02 $+$ & 7.71E+02 $+$ & 6.03E+03 $+$ & 4.35E+02 \\
\bottomrule
	\end{tabular}
}
\end{table*}

\begin{table*}
	\renewcommand{\arraystretch}{1.0}
	\caption{Average results obtained by ECI-BO and the compared SAEAs  on 30-D CEC 2013 test problems}
	\label{table_SAEA_CEC2013_30D}
	\centering
	\resizebox{\textwidth}{!}{
		\begin{tabular}{c c c c c c c  c c } 
			\toprule
			$f$   & $d$& GPDE &  GPEME & MGP-SLPSO  &  IKAEA &  KAEA-AEI & L2SMEA &  ECI-BO   \\
			\midrule
			$f_{1}$ & 30 & -1.02E+03 $+$ & 4.78E+04 $+$ & 2.24E+03 $+$ & -1.40E+03 $+$ & -1.23E+03 $+$ & 1.23E+04 $+$ & -1.40E+03 \\
			$f_{2}$ & 30 & 1.38E+08 $+$ & 9.89E+08 $+$ & 1.29E+08 $+$ & 6.18E+07 $+$ & 2.15E+08 $+$ & 2.73E+08 $+$ & 3.42E+07 \\
			$f_{3}$ & 30 & 1.32E+18 $+$ & 1.98E+18 $+$ & -8.34E+26 $-$ & 1.48E+11 $\approx$ & 1.08E+11 $-$ & 1.46E+11 $\approx$ & 6.15E+11 \\
			$f_{4}$ & 30 & 1.38E+05 $\approx$ & 1.42E+05 $\approx$ & 1.35E+05 $-$ & 1.55E+05 $\approx$ & 1.27E+05 $-$ & 1.50E+05 $\approx$ & 1.54E+05 \\
			$f_{5}$ & 30 & 8.18E+03 $+$ & 2.46E+04 $+$ & -1.94E+02 $+$ & 6.98E+01 $+$ & 3.37E+03 $+$ & 9.26E+03 $+$ & -5.37E+02 \\
			$f_{6}$ & 30 & 3.73E+02 $+$ & 7.73E+03 $+$ & -3.31E+02 $+$ & -7.55E+02 $+$ & -2.11E+02 $+$ & 2.64E+02 $+$ & -8.06E+02 \\
			$f_{7}$ & 30 & 6.23E+05 $+$ & 4.78E+05 $+$ & 2.12E+03 $+$ & -3.45E+02 $\approx$ & -4.54E+02 $+$ & -5.35E+02 $-$ & -4.78E+02 \\
\bottomrule
		\end{tabular}
	}
\end{table*}

\begin{table*}
	\renewcommand{\arraystretch}{1.0}
	\caption{Average results obtained by ECI-BO and the compared SAEAs  on 50-D CEC 2013 test problems}
	\label{table_SAEA_CEC2013_50D}
	\centering
	\resizebox{\textwidth}{!}{
		\begin{tabular}{c c c c c c c  c c } 
			\toprule
			$f$   & $d$&GPDE &  GPEME & MGP-SLPSO  &  IKAEA &  KAEA-AEI & L2SMEA &  ECI-BO   \\
			\midrule
			$f_{1}$ & 50 & -1.23E+02 $+$ & -1.27E+03 $+$ & -1.40E+03 $-$ & -1.37E+03 $+$ & -6.60E+02 $+$ & 3.36E+04 $+$ & -1.40E+03 \\
			$f_{2}$ & 50 & 3.91E+08 $+$ & 1.33E+08 $+$ & 1.69E+08 $+$ & 2.45E+08 $+$ & 2.25E+08 $+$ & 4.97E+08 $+$ & 8.66E+07 \\
			$f_{3}$ & 50 & 7.12E+16 $+$ & 3.33E+14 $+$ & 6.54E+15 $+$ & 6.62E+15 $\approx$ & 5.70E+16 $+$ & 1.21E+12 $\approx$ & 1.28E+12 \\
			$f_{4}$ & 50 & 2.16E+05 $-$ & 2.04E+05 $-$ & 1.81E+05 $-$ & 2.22E+05 $-$ & 2.05E+05 $-$ & 1.06E+05 $-$ & 2.41E+05 \\
			$f_{5}$ & 50 & 4.52E+03 $+$ & 6.68E+03 $+$ & 4.25E+03 $+$ & 3.41E+03 $+$ & 3.84E+03 $+$ & 9.24E+03 $+$ & 2.48E+03 \\
			$f_{6}$ & 50 & -3.50E+02 $+$ & -7.97E+02 $-$ & -5.22E+02 $+$ & -5.04E+02 $+$ & -4.15E+02 $+$ & 1.20E+03 $+$ & -7.35E+02 \\
			$f_{7}$ & 50 & 1.25E+05 $+$ & 1.37E+04 $+$ & 6.51E+04 $+$ & 4.39E+04 $+$ & 1.02E+05 $+$ & -3.13E+02 $\approx$ & 3.21E+02 \\
\bottomrule
		\end{tabular}
	}
\end{table*}

\begin{table*}
	\renewcommand{\arraystretch}{1.0}
	\caption{Average results obtained by ECI-BO and the compared SAEAs  on 100-D CEC 2013 test problems}
	\label{table_SAEA_CEC2013_100D}
	\centering
	\resizebox{\textwidth}{!}{
		\begin{tabular}{c c c c c c c  c c } 
			\toprule
			$f$   & $d$& GPDE &  GPEME & MGP-SLPSO  &  IKAEA &  KAEA-AEI & L2SMEA &  ECI-BO   \\
			\midrule
			$f_{1}$ & 100 & 2.15E+04 $+$ & 1.10E+04 $+$ & -1.40E+03 $+$ & 3.82E+03 $+$ & 1.81E+04 $+$ & 6.63E+04 $+$ & -1.40E+03 \\
			$f_{2}$ & 100 & 1.10E+09 $+$ & 4.74E+08 $+$ & 1.14E+09 $+$ & 7.09E+08 $+$ & 8.41E+08 $+$ & 1.24E+09 $+$ & 3.48E+08 \\
			$f_{3}$ & 100 & 5.53E+23 $+$ & 6.60E+21 $+$ & 3.25E+23 $+$ & 2.95E+22 $\approx$ & 6.71E+23 $+$ & 2.86E+17 $-$ & 5.88E+19 \\
			$f_{4}$ & 100 & 4.54E+05 $-$ & 4.34E+05 $-$ & 3.71E+05 $-$ & 4.98E+05 $-$ & 4.55E+05 $-$ & 3.14E+05 $-$ & 5.23E+05 \\
			$f_{5}$ & 100 & 1.59E+04 $-$ & 1.11E+04 $-$ & 2.72E+04 $-$ & 1.55E+04 $-$ & 1.48E+04 $-$ & 1.61E+04 $-$ & 4.71E+04 \\
			$f_{6}$ & 100 & 3.09E+03 $+$ & 1.29E+03 $+$ & 2.54E+03 $+$ & 1.59E+03 $+$ & 2.89E+03 $+$ & 9.50E+03 $+$ & -9.13E+01 \\
			$f_{7}$ & 100 & 3.17E+08 $+$ & 4.47E+07 $+$ & 3.77E+08 $+$ & 4.28E+07 $+$ & 4.47E+08 $+$ & 1.35E+05 $-$ & 9.48E+05 \\
\bottomrule
		\end{tabular}
	}
\end{table*}

\begin{table*}
	\renewcommand{\arraystretch}{1.0}
	\caption{Average results obtained by ECI-BO and the compared SAEAs on 30-D CEC 2017 test porblems}
	\label{table_SAEA_CEC2017_30D}
	\centering
	\resizebox{\textwidth}{!}{
		\begin{tabular}{c c c c c c c  c c } 
			\toprule
			$f$   & $d$& GPDE &  GPEME & MGP-SLPSO  &  IKAEA &  KAEA-AEI & L2SMEA &  ECI-BO   \\
			\midrule
			$f_{1}$ & 30 & 1.44E+08 $+$ & 1.94E+06 $+$ & 1.89E+08 $+$ & 8.87E+05 $+$ & 4.38E+07 $+$ & 2.56E+10 $+$ & 1.06E+04 \\
			$f_{3}$ & 30 & 2.14E+05 $-$ & 2.01E+05 $-$ & 1.71E+05 $-$ & 2.02E+05 $-$ & 2.13E+05 $-$ & 1.09E+05 $-$ & 2.74E+05 \\
			$f_{4}$ & 30 & 1.11E+03 $+$ & 5.24E+02 $-$ & 1.06E+03 $+$ & 9.65E+02 $+$ & 8.93E+02 $+$ & 5.36E+03 $+$ & 5.80E+02 \\
			$f_{5}$ & 30 & 7.63E+02 $+$ & 5.80E+02 $-$ & 5.54E+02 $-$ & 7.84E+02 $+$ & 7.45E+02 $+$ & 8.40E+02 $+$ & 6.66E+02 \\
			$f_{6}$ & 30 & 6.19E+02 $+$ & 6.15E+02 $+$ & 6.12E+02 $+$ & 6.31E+02 $+$ & 6.19E+02 $+$ & 6.74E+02 $+$ & 6.06E+02 \\
			$f_{7}$ & 30 & 1.03E+03 $+$ & 9.89E+02 $+$ & 8.35E+02 $-$ & 1.03E+03 $+$ & 1.02E+03 $+$ & 1.46E+03 $+$ & 9.33E+02 \\
			$f_{8}$ & 30 & 1.06E+03 $+$ & 8.77E+02 $-$ & 8.54E+02 $-$ & 1.08E+03 $+$ & 1.03E+03 $+$ & 1.12E+03 $+$ & 9.45E+02 \\
\bottomrule
		\end{tabular}
	}
\end{table*}

\begin{table*}
	\renewcommand{\arraystretch}{1.0}
	\caption{Average results obtained by ECI-BO and the compared SAEAs on 50-D CEC 2017 test porblems}
	\label{table_SAEA_CEC2017_50D}
	\centering
	\resizebox{\textwidth}{!}{
		\begin{tabular}{c c c c c c c  c c } 
			\toprule
			$f$   & $d$& GPDE &  GPEME & MGP-SLPSO  &  IKAEA &  KAEA-AEI & L2SMEA &  ECI-BO   \\
			\midrule
		$f_{1}$ & 50 & 2.47E+09 $+$ & 2.78E+08 $+$ & 1.57E+09 $+$ & 1.11E+08 $+$ & 1.45E+09 $+$ & 5.20E+10 $+$ & 3.73E+05 \\
		$f_{3}$ & 50 & 4.24E+05 $-$ & 5.25E+05 $\approx$ & 3.58E+05 $-$ & 4.33E+05 $-$ & 4.92E+05 $+$ & 2.11E+05 $-$ & 4.89E+05 \\
		$f_{4}$ & 50 & 2.25E+03 $+$ & 8.05E+02 $\approx$ & 1.92E+03 $+$ & 1.74E+03 $+$ & 2.06E+03 $+$ & 1.04E+04 $+$ & 8.02E+02 \\
		$f_{5}$ & 50 & 9.94E+02 $+$ & 7.15E+02 $-$ & 6.37E+02 $-$ & 1.04E+03 $+$ & 9.86E+02 $+$ & 1.11E+03 $+$ & 7.88E+02 \\
		$f_{6}$ & 50 & 6.34E+02 $+$ & 6.37E+02 $+$ & 6.21E+02 $+$ & 6.41E+02 $+$ & 6.35E+02 $+$ & 6.77E+02 $+$ & 6.06E+02 \\
		$f_{7}$ & 50 & 1.35E+03 $+$ & 1.35E+03 $+$ & 9.62E+02 $-$ & 1.32E+03 $+$ & 1.32E+03 $+$ & 2.16E+03 $+$ & 1.15E+03 \\
		$f_{8}$ & 50 & 1.30E+03 $+$ & 1.02E+03 $-$ & 9.53E+02 $-$ & 1.33E+03 $+$ & 1.29E+03 $+$ & 1.40E+03 $+$ & 1.07E+03 \\
\bottomrule
		\end{tabular}
	}
\end{table*}

\begin{table*}
	\renewcommand{\arraystretch}{1.0}
	\caption{Average results obtained by ECI-BO and the compared SAEAs on 100-D CEC 2017 test porblems}
	\label{table_SAEA_CEC2017_100D}
	\centering
	\resizebox{\textwidth}{!}{
		\begin{tabular}{c c c c c c c  c c } 
			\toprule
			$f$   & $d$& GPDE &  GPEME & MGP-SLPSO  &  IKAEA &  KAEA-AEI & L2SMEA &  ECI-BO   \\
			\midrule
		$f_{1}$ & 100 & 4.05E+10 $+$ & 2.40E+10 $+$ & 9.75E+09 $+$ & 1.10E+10 $+$ & 3.51E+10 $+$ & 1.22E+11 $+$ & 2.76E+07 \\
		$f_{3}$ & 100 & 2.34E+06 $\approx$ & 1.84E+07 $+$ & 3.08E+06 $\approx$ & 2.10E+06 $\approx$ & 8.27E+06 $+$ & 5.05E+05 $-$ & 9.70E+05 \\
		$f_{4}$ & 100 & 9.21E+03 $+$ & 4.43E+03 $+$ & 1.36E+04 $+$ & 6.69E+03 $+$ & 8.16E+03 $+$ & 1.82E+04 $+$ & 3.37E+03 \\
		$f_{5}$ & 100 & 1.70E+03 $+$ & 1.40E+03 $\approx$ & 1.43E+03 $\approx$ & 1.73E+03 $+$ & 1.69E+03 $+$ & 1.92E+03 $+$ & 1.42E+03 \\
		$f_{6}$ & 100 & 6.61E+02 $+$ & 6.59E+02 $+$ & 6.47E+02 $+$ & 6.68E+02 $+$ & 6.66E+02 $+$ & 6.88E+02 $+$ & 6.12E+02 \\
		$f_{7}$ & 100 & 2.59E+03 $+$ & 2.86E+03 $+$ & 1.60E+03 $-$ & 2.44E+03 $+$ & 2.48E+03 $+$ & 4.44E+03 $+$ & 1.81E+03 \\
		$f_{8}$ & 100 & 2.04E+03 $+$ & 1.68E+03 $\approx$ & 1.71E+03 $\approx$ & 2.05E+03 $+$ & 1.99E+03 $+$ & 2.24E+03 $+$ & 1.71E+03 \\
\bottomrule
		\end{tabular}
	}
\end{table*}

\begin{figure*}	
	\centering
	\makebox[\textwidth]{\parbox{1.2\textwidth}{
			\subfloat[30-D Ellipsoid]{\includegraphics[width=0.24\linewidth]{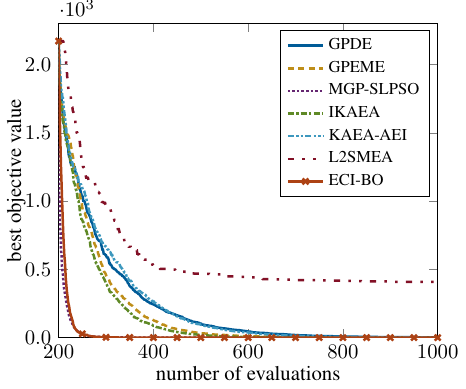}} 	\hfil
			\subfloat[50-D Ellipsoid]{\includegraphics[width=0.24\linewidth]{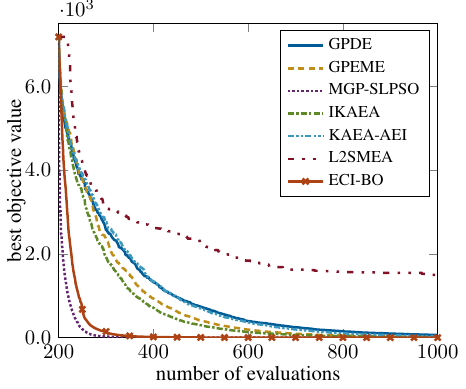}}\hfil
			\subfloat[100-D Ellipsoid]{\includegraphics[width=0.24\linewidth]{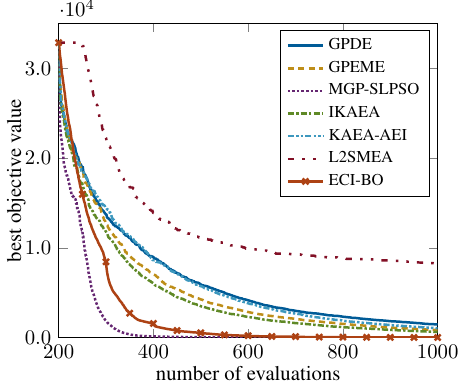}}\hfil
			\subfloat[200-D Ellipsoid]{\includegraphics[width=0.24\linewidth]{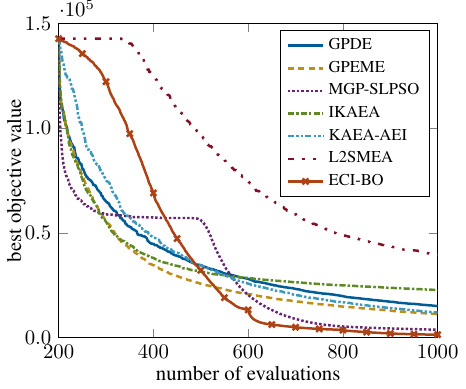}}   \vspace{-3mm} \\
			\subfloat[30-D Rosenbrock]{\includegraphics[width=0.24\linewidth]{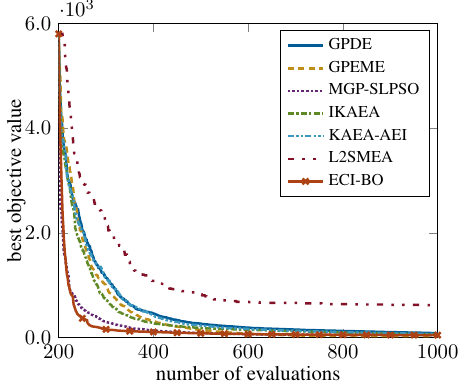}} 	\hfil
			\subfloat[50-D Rosenbrock]{\includegraphics[width=0.24\linewidth]{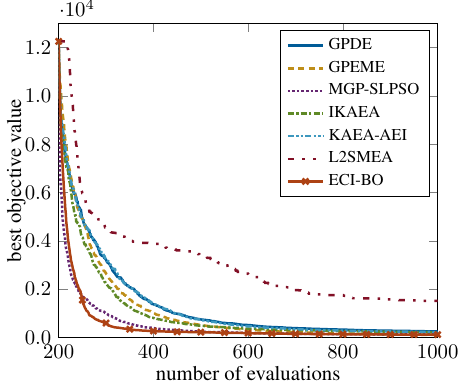}}\hfil
			\subfloat[100-D Rosenbrock]{\includegraphics[width=0.24\linewidth]{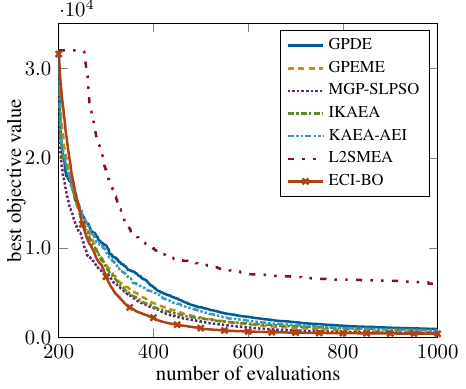}}\hfil
			\subfloat[200-D Rosenbrock]{\includegraphics[width=0.24\linewidth]{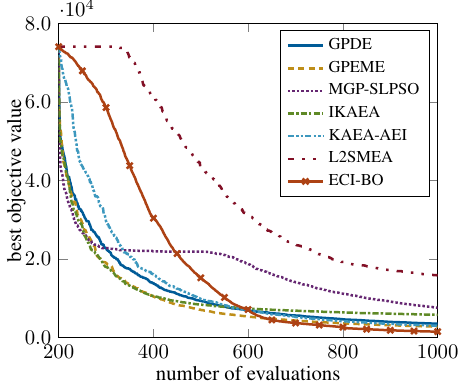}} \vspace{-3mm}  \\ 
			\subfloat[30-D Ackley]{\includegraphics[width=0.24\linewidth]{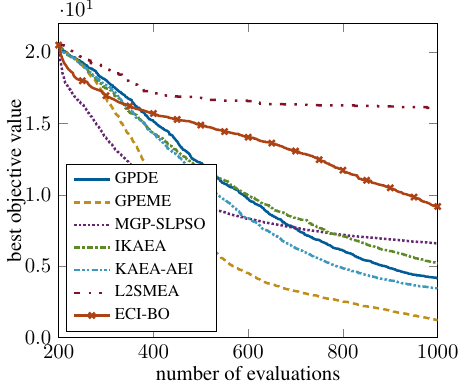}} 	\hfil
			\subfloat[50-D Ackley]{\includegraphics[width=0.24\linewidth]{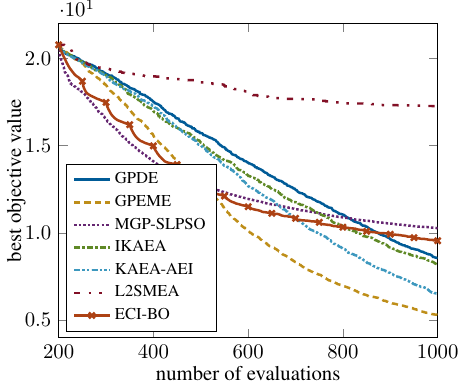}}\hfil
			\subfloat[100-D Ackley]{\includegraphics[width=0.24\linewidth]{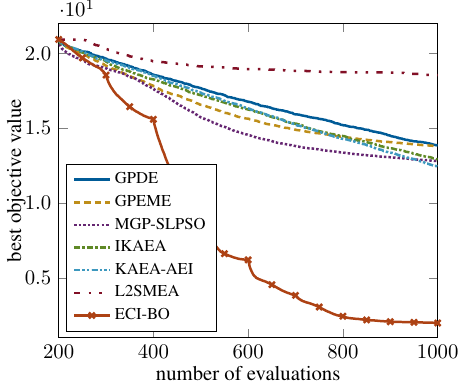}}\hfil
			\subfloat[200-D Ackley]{\includegraphics[width=0.24\linewidth]{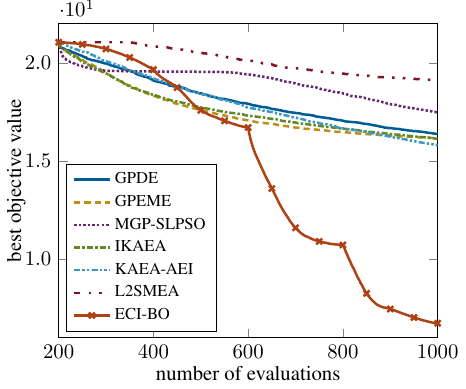}} \vspace{-3mm}  \\ 
			\subfloat[30-D Griewank]{\includegraphics[width=0.24\linewidth]{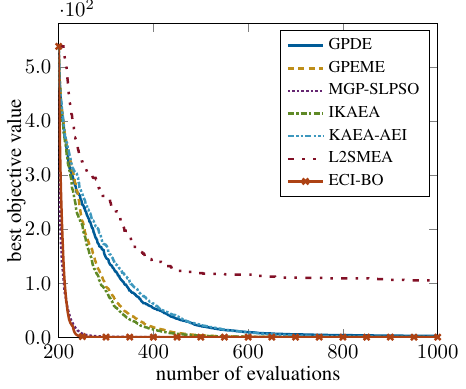}} 	\hfil
			\subfloat[50-D Griewank]{\includegraphics[width=0.24\linewidth]{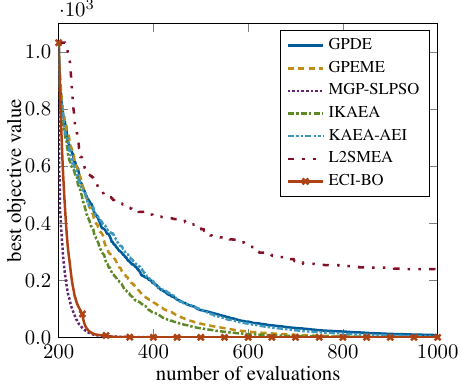}}\hfil
			\subfloat[100-D Griewank]{\includegraphics[width=0.24\linewidth]{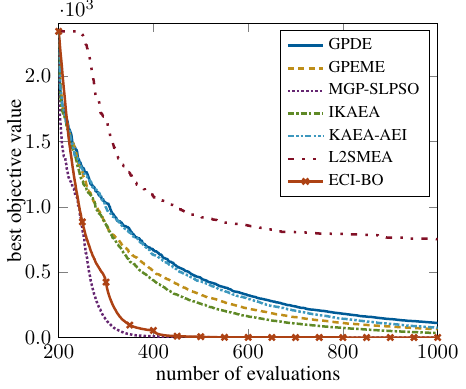}}\hfil
			\subfloat[200-D Griewank]{\includegraphics[width=0.24\linewidth]{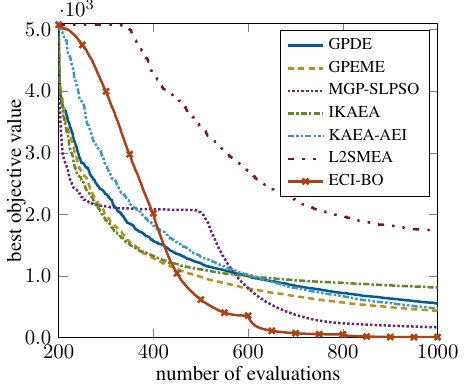}}  \vspace{-3mm}  \\ 
			\subfloat[30-D Rastrigin]{\includegraphics[width=0.24\linewidth]{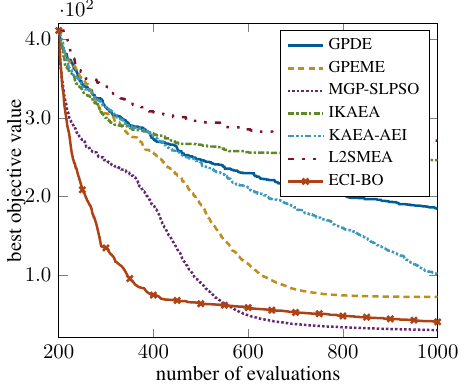}} 	\hfil
			\subfloat[50-D Rastrigin]{\includegraphics[width=0.24\linewidth]{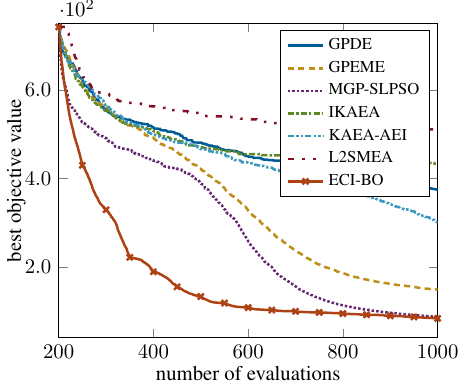}}\hfil
			\subfloat[100-D Rastrigin]{\includegraphics[width=0.24\linewidth]{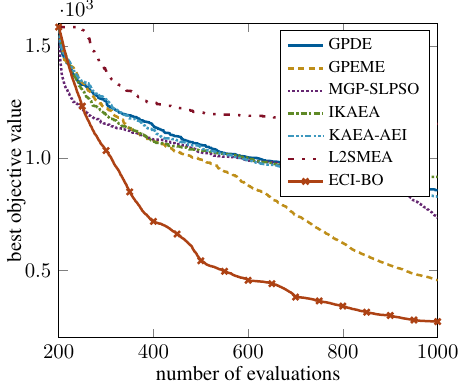}}\hfil
			\subfloat[200-D Rastrigin]{\includegraphics[width=0.24\linewidth]{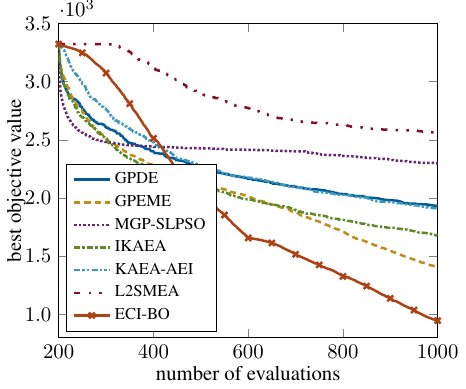}} \vspace{-3mm}  \\ 
	}}
	\caption{Convergence plots of the compared SAEAs on the analytical test problems.}
	\label{fig_comparison_SAEA}
\end{figure*}

The experiment results are given in Tables~\ref{table_SAEA_analytical} to~\ref{table_SAEA_CEC2017_100D}, where the average results of 30 runs are shown.  We also conduct the Wilcoxon signed rank test with significance level of $\alpha=0.05$, and  use $+$, $-$ and $\approx$ to donate that the results of the proposed ECI-BO are significantly better than, worse than and similar to the results of the compared algorithms, respectively.   From the tables we can find that our proposed ECI-BO performs significantly better than the compared SAEAs on most of the test problems in terms of the final optimization results. It finds significantly better results on most of the benchmark problems than the compared algorithms.  The SAEAs employ a EA population to search in the high-dimensional space and utilize the GP to select promising individuals for expensive evaluation. In comparison, our proposed ECI-BO finds solutions for expensive evaluation by searching along a one-dimensional space, which is much easier to search than in the original high-dimensional space. Through iterating over different coordinates, the proposed approach is able to improve the solution in the original space gradually.

The convergence plots of the compared algorithms are given in Fig.~\ref{fig_comparison_SAEA}, where the average values of 30 runs are plotted. First, we can find that the L2SMEA approach finds the worst results on most of the test problems. The GP models that the L2SMEA trains are on one-dimensional spaces. Therefore, the accuracy of these GP models are often poor. This might be the reason that the L2SMEA performs worse than the other approaches. The L2SMEA approach might be more suitable to be used in large-scale problems.   Compared with other algorithms, the proposed ECI-BO is able to find very competitive results on most of the test problems, especially on 100-D and 200-D test problems. On the 30-D and 50-D Ackley problems, the proposed ECI-BO is outperformed by the compared algorithms. But on the 100-D and 200-D Ackley problems, the proposed ECI-BO is able to find the best optimization results at the end of 1000 evaluations. This shows the advantage of the proposed ECI-BO on solving high-dimensional optimization  problems.

\section{Conclusions}
\label{section_conclusion}

In this work, we propose a simple and efficient approach to extend the Bayesian optimization algorithm to high-dimensional problems. We propose the expected coordinate improvement (ECI) criterion to measure the amount of improvement when moving the current best solution along one coordinate. Based on the ECI, we propose a high-dimensional Bayesian optimization approach which optimizes one coordinate at each iteration following the order of the coordinates' ECI values. The proposed algorithm shows very competitive performance on the selected test suite when compared with the standard Bayesian optimization algorithm and the state-of-the-art high-dimensional approaches.  Our work provides an easy-to-understand and simple-to-implement approach for the Bayesian optimization field and has the potential to be used in solving real-world high-dimensional expensive optimization problems. However, our work does not consider expensive constraints and multiple expensive objectives. Applying the expected coordinate improvement approach to constrained optimization and multiobjective optimization would be very interesting work for future research.

\section{Acknowledgments}
This work was supported by the National Natural Science Foundation of China under Grant 62106207.




 \bibliographystyle{elsarticle-num-names} 
\bibliography{My_Reference.bib}

\begin{thebibliography}{64}
\expandafter\ifx\csname natexlab\endcsname\relax\def\natexlab#1{#1}\fi
\providecommand{\url}[1]{\texttt{#1}}
\providecommand{\href}[2]{#2}
\providecommand{\path}[1]{#1}
\providecommand{\DOIprefix}{doi:}
\providecommand{\ArXivprefix}{arXiv:}
\providecommand{\URLprefix}{URL: }
\providecommand{\Pubmedprefix}{pmid:}
\providecommand{\doi}[1]{\href{http://dx.doi.org/#1}{\path{#1}}}
\providecommand{\Pubmed}[1]{\href{pmid:#1}{\path{#1}}}
\providecommand{\bibinfo}[2]{#2}
\ifx\xfnm\relax \def\xfnm[#1]{\unskip,\space#1}\fi
\bibitem[{Mockus(1974)}]{Mockus_1975}
\bibinfo{author}{J.~Mockus},
\newblock \bibinfo{title}{On bayesian methods for seeking the extremum},
\newblock in: \bibinfo{booktitle}{Proceedings of the IFIP Technical
  Conference}, \bibinfo{year}{1974}, pp. \bibinfo{pages}{400--404}.
\bibitem[{Mockus(1994)}]{Mockus_1994}
\bibinfo{author}{J.~Mockus},
\newblock \bibinfo{title}{Application of bayesian approach to numerical methods
  of global and stochastic optimization},
\newblock \bibinfo{journal}{Journal of Global Optimization} \bibinfo{volume}{4}
  (\bibinfo{year}{1994}) \bibinfo{pages}{347--365}.
  \DOIprefix\doi{https://doi.org/10.1007/bf01099263}.
\bibitem[{Jones et~al.(1998)Jones, Schonlau, and Welch}]{Jones_1998}
\bibinfo{author}{D.~R. Jones}, \bibinfo{author}{M.~Schonlau},
  \bibinfo{author}{W.~J. Welch},
\newblock \bibinfo{title}{Efficient global optimization of expensive black-box
  functions},
\newblock \bibinfo{journal}{Journal of Global Optimization}
  \bibinfo{volume}{13} (\bibinfo{year}{1998}) \bibinfo{pages}{455--492}.
  \DOIprefix\doi{https://doi.org/10.1023/A:1008306431147}.
\bibitem[{Rasmussen and Williams(2006)}]{Rasmussen_2006}
\bibinfo{author}{C.~E. Rasmussen}, \bibinfo{author}{C.~K. Williams},
  \bibinfo{title}{Gaussian processes for machine learning},
  \bibinfo{publisher}{The MIT Press, Cambridge, MA, USA}, \bibinfo{year}{2006}.
\bibitem[{Jones(2001)}]{Jones_2001}
\bibinfo{author}{D.~R. Jones},
\newblock \bibinfo{title}{A taxonomy of global optimization methods based on
  response surfaces},
\newblock \bibinfo{journal}{Journal of Global Optimization}
  \bibinfo{volume}{21} (\bibinfo{year}{2001}) \bibinfo{pages}{345--383}.
  \DOIprefix\doi{https://doi.org/10.1023/A:1012771025575}.
\bibitem[{Snoek et~al.(2012)Snoek, Larochelle, and Adams}]{Snoek_2012}
\bibinfo{author}{J.~Snoek}, \bibinfo{author}{H.~Larochelle},
  \bibinfo{author}{R.~P. Adams},
\newblock \bibinfo{title}{Practical {B}ayesian optimization of machine learning
  algorithms},
\newblock in: \bibinfo{booktitle}{Advances in Neural Information Processing
  Systems}, \bibinfo{year}{2012}, pp. \bibinfo{pages}{2951--2959}.
\bibitem[{Shahriari et~al.(2016)Shahriari, Swersky, Wang, Adams, and
  Freitas}]{Shahriari_2016}
\bibinfo{author}{B.~Shahriari}, \bibinfo{author}{K.~Swersky},
  \bibinfo{author}{Z.~Wang}, \bibinfo{author}{R.~P. Adams},
  \bibinfo{author}{N.~d. Freitas},
\newblock \bibinfo{title}{Taking the human out of the loop: A review of
  bayesian optimization},
\newblock \bibinfo{journal}{Proceedings of the IEEE} \bibinfo{volume}{104}
  (\bibinfo{year}{2016}) \bibinfo{pages}{148--175}.
  \DOIprefix\doi{https://doi.org/10.1109/JPROC.2015.2494218}.
\bibitem[{Guo et~al.(2021)Guo, Ong, and Liu}]{Guo_2021}
\bibinfo{author}{Z.~Guo}, \bibinfo{author}{Y.-S. Ong},
  \bibinfo{author}{H.~Liu},
\newblock \bibinfo{title}{Calibrated and recalibrated expected improvements for
  bayesian optimization},
\newblock \bibinfo{journal}{Structural and Multidisciplinary Optimization}
  \bibinfo{volume}{64} (\bibinfo{year}{2021}) \bibinfo{pages}{3549--3567}.
  \DOIprefix\doi{10.1007/s00158-021-03038-3}.
\bibitem[{Zhang et~al.(2010)Zhang, Liu, Tsang, and Virginas}]{Zhang_2010}
\bibinfo{author}{Q.~Zhang}, \bibinfo{author}{W.~Liu},
  \bibinfo{author}{E.~Tsang}, \bibinfo{author}{B.~Virginas},
\newblock \bibinfo{title}{Expensive multiobjective optimization by {MOEA/D}
  with {G}aussian process model},
\newblock \bibinfo{journal}{IEEE Transactions on Evolutionary Computation}
  \bibinfo{volume}{14} (\bibinfo{year}{2010}) \bibinfo{pages}{456--474}.
  \DOIprefix\doi{https://doi.org/10.1109/TEVC.2009.2033671}.
\bibitem[{Yang et~al.(2019)Yang, Emmerich, Deutz, and Bäck}]{Yang_2019}
\bibinfo{author}{K.~Yang}, \bibinfo{author}{M.~Emmerich},
  \bibinfo{author}{A.~Deutz}, \bibinfo{author}{T.~Bäck},
\newblock \bibinfo{title}{Multi-objective bayesian global optimization using
  expected hypervolume improvement gradient},
\newblock \bibinfo{journal}{Swarm and Evolutionary Computation}
  \bibinfo{volume}{44} (\bibinfo{year}{2019}) \bibinfo{pages}{945--956}.
  \DOIprefix\doi{https://doi.org/10.1016/j.swevo.2018.10.007}.
\bibitem[{Han and Ouyang(2022)}]{Han_2022}
\bibinfo{author}{M.~Han}, \bibinfo{author}{L.~Ouyang},
\newblock \bibinfo{title}{A novel bayesian approach for multi-objective
  stochastic simulation optimization},
\newblock \bibinfo{journal}{Swarm and Evolutionary Computation}
  \bibinfo{volume}{75} (\bibinfo{year}{2022}) \bibinfo{pages}{101192}.
  \DOIprefix\doi{https://doi.org/10.1016/j.swevo.2022.101192}.
\bibitem[{Basudhar et~al.(2012)Basudhar, Dribusch, Lacaze, and
  Missoum}]{Basudhar_2012}
\bibinfo{author}{A.~Basudhar}, \bibinfo{author}{C.~Dribusch},
  \bibinfo{author}{S.~Lacaze}, \bibinfo{author}{S.~Missoum},
\newblock \bibinfo{title}{Constrained efficient global optimization with
  support vector machines},
\newblock \bibinfo{journal}{Structural and Multidisciplinary Optimization}
  \bibinfo{volume}{46} (\bibinfo{year}{2012}) \bibinfo{pages}{201--221}.
  \DOIprefix\doi{10.1007/s00158-011-0745-5}.
\bibitem[{Li and Zhang(2021)}]{LiGH_2021}
\bibinfo{author}{G.~Li}, \bibinfo{author}{Q.~Zhang},
\newblock \bibinfo{title}{Multiple penalties and multiple local surrogates for
  expensive constrained optimization},
\newblock \bibinfo{journal}{IEEE Transactions on Evolutionary Computation}
  \bibinfo{volume}{25} (\bibinfo{year}{2021}) \bibinfo{pages}{769--778}.
  \DOIprefix\doi{10.1109/TEVC.2021.3066606}.
\bibitem[{Briffoteaux et~al.(2020)Briffoteaux, Gobert, Ragonnet, Gmys, Mezmaz,
  Melab, and Tuyttens}]{Briffoteaux_2020}
\bibinfo{author}{G.~Briffoteaux}, \bibinfo{author}{M.~Gobert},
  \bibinfo{author}{R.~Ragonnet}, \bibinfo{author}{J.~Gmys},
  \bibinfo{author}{M.~Mezmaz}, \bibinfo{author}{N.~Melab},
  \bibinfo{author}{D.~Tuyttens},
\newblock \bibinfo{title}{Parallel surrogate-assisted optimization: Batched
  bayesian neural network-assisted ga versus q-ego},
\newblock \bibinfo{journal}{Swarm and Evolutionary Computation}
  \bibinfo{volume}{57} (\bibinfo{year}{2020}) \bibinfo{pages}{100717}.
  \DOIprefix\doi{https://doi.org/10.1016/j.swevo.2020.100717}.
\bibitem[{Chen et~al.(2023)Chen, Luo, Li, and Wang}]{Chen_2023}
\bibinfo{author}{J.~Chen}, \bibinfo{author}{F.~Luo}, \bibinfo{author}{G.~Li},
  \bibinfo{author}{Z.~Wang},
\newblock \bibinfo{title}{Batch bayesian optimization with adaptive batch
  acquisition functions via multi-objective optimization},
\newblock \bibinfo{journal}{Swarm and Evolutionary Computation}
  \bibinfo{volume}{79} (\bibinfo{year}{2023}) \bibinfo{pages}{101293}.
  \DOIprefix\doi{https://doi.org/10.1016/j.swevo.2023.101293}.
\bibitem[{Zhan et~al.(2023)Zhan, Meng, and Xing}]{Zhan_2023}
\bibinfo{author}{D.~Zhan}, \bibinfo{author}{Y.~Meng},
  \bibinfo{author}{H.~Xing},
\newblock \bibinfo{title}{A fast multipoint expected improvement for parallel
  expensive optimization},
\newblock \bibinfo{journal}{IEEE Transactions on Evolutionary Computation}
  \bibinfo{volume}{27} (\bibinfo{year}{2023}) \bibinfo{pages}{170--184}.
  \DOIprefix\doi{https://doi.org/10.1109/TEVC.2022.3168060}.
\bibitem[{Wang et~al.(2023)Wang, Zhang, Ong, Yao, Liu, and Luo}]{WangZ_2023}
\bibinfo{author}{Z.~Wang}, \bibinfo{author}{Q.~Zhang}, \bibinfo{author}{Y.~S.
  Ong}, \bibinfo{author}{S.~Yao}, \bibinfo{author}{H.~Liu},
  \bibinfo{author}{J.~Luo},
\newblock \bibinfo{title}{Choose appropriate subproblems for collaborative
  modeling in expensive multiobjective optimization},
\newblock \bibinfo{journal}{IEEE Transactions on Cybernetics}
  \bibinfo{volume}{53} (\bibinfo{year}{2023}) \bibinfo{pages}{483--496}.
  \DOIprefix\doi{10.1109/TCYB.2021.3126341}.
\bibitem[{Binois and Wycoff(2022)}]{Binois_2022}
\bibinfo{author}{M.~Binois}, \bibinfo{author}{N.~Wycoff},
\newblock \bibinfo{title}{A survey on high-dimensional {G}aussian process
  modeling with application to {B}ayesian optimization},
\newblock \bibinfo{journal}{ACM Transactions on Evolutionary Learning and
  Optimization} \bibinfo{volume}{2} (\bibinfo{year}{2022})
  \bibinfo{pages}{1--26}. \DOIprefix\doi{https://doi.org/10.1145/3545611}.
\bibitem[{Linkletter et~al.(2006)Linkletter, Bingham, Hengartner, Higdon, and
  Ye}]{Linkletter_2006}
\bibinfo{author}{C.~Linkletter}, \bibinfo{author}{D.~Bingham},
  \bibinfo{author}{N.~Hengartner}, \bibinfo{author}{D.~Higdon},
  \bibinfo{author}{K.~Q. Ye},
\newblock \bibinfo{title}{Variable selection for gaussian process models in
  computer experiments},
\newblock \bibinfo{journal}{Technometrics} \bibinfo{volume}{48}
  (\bibinfo{year}{2006}) \bibinfo{pages}{478--490}.
  \DOIprefix\doi{https://doi.org/10.1198/004017006000000228}.
\bibitem[{Chen et~al.(2012)Chen, Castro, and Krause}]{Chen_2012}
\bibinfo{author}{B.~Chen}, \bibinfo{author}{R.~M. Castro},
  \bibinfo{author}{A.~Krause},
\newblock \bibinfo{title}{Joint optimization and variable selection of
  high-dimensional {G}aussian processes},
\newblock in: \bibinfo{booktitle}{International Conference on Machine
  Learning}, \bibinfo{year}{2012}, p. \bibinfo{pages}{1379–1386}.
\bibitem[{Zhang et~al.(2023)Zhang, Chen, Hung, and Deng}]{Zhang_2023}
\bibinfo{author}{F.~Zhang}, \bibinfo{author}{R.-B. Chen},
  \bibinfo{author}{Y.~Hung}, \bibinfo{author}{X.~Deng},
\newblock \bibinfo{title}{Indicator-based {B}ayesian variable selection for
  {G}aussian process models in computer experiments},
\newblock \bibinfo{journal}{Computational Statistics \& Data Analysis}
  \bibinfo{volume}{185} (\bibinfo{year}{2023}) \bibinfo{pages}{107757}.
  \DOIprefix\doi{https://doi.org/10.1016/j.csda.2023.107757}.
\bibitem[{Ulmasov et~al.(2016)Ulmasov, Baroukh, Chachuat, Deisenroth, and
  Misener}]{Ulmasov_2016}
\bibinfo{author}{D.~Ulmasov}, \bibinfo{author}{C.~Baroukh},
  \bibinfo{author}{B.~Chachuat}, \bibinfo{author}{M.~P. Deisenroth},
  \bibinfo{author}{R.~Misener},
\newblock \bibinfo{title}{Bayesian optimization with dimension scheduling:
  Application to biological systems},
\newblock \bibinfo{journal}{Computer Aided Chemical Engineering}
  \bibinfo{volume}{38} (\bibinfo{year}{2016}) \bibinfo{pages}{1051--1056}.
  \DOIprefix\doi{https://doi.org/10.1016/B978-0-444-63428-3.50180-6}.
\bibitem[{Li et~al.(2017)Li, Gupta, Rana, Nguyen, Venkatesh, and
  Shilton}]{Li_2017}
\bibinfo{author}{C.~Li}, \bibinfo{author}{S.~Gupta}, \bibinfo{author}{S.~Rana},
  \bibinfo{author}{T.~V. Nguyen}, \bibinfo{author}{S.~Venkatesh},
  \bibinfo{author}{A.~Shilton},
\newblock \bibinfo{title}{High dimensional {B}ayesian optimization using
  dropout},
\newblock in: \bibinfo{booktitle}{International Joint Conference on Artificial
  Intelligence}, \bibinfo{year}{2017}, pp. \bibinfo{pages}{2096--2102}.
\bibitem[{Song et~al.(2022)Song, Xue, Huang, and Qian}]{Song_2022}
\bibinfo{author}{L.~Song}, \bibinfo{author}{K.~Xue},
  \bibinfo{author}{X.~Huang}, \bibinfo{author}{C.~Qian},
\newblock \bibinfo{title}{Monte carlo tree search based variable selection for
  high dimensional bayesian optimization},
\newblock in: \bibinfo{booktitle}{Advances in Neural Information Processing
  Systems}, volume~\bibinfo{volume}{35}, \bibinfo{year}{2022}, pp.
  \bibinfo{pages}{28488--28501}.
\bibitem[{Wang et~al.(2016)Wang, Hutter, Zoghi, Matheson, and
  de~Feitas}]{Wang_2016}
\bibinfo{author}{Z.~Wang}, \bibinfo{author}{F.~Hutter},
  \bibinfo{author}{M.~Zoghi}, \bibinfo{author}{D.~Matheson},
  \bibinfo{author}{N.~de~Feitas},
\newblock \bibinfo{title}{Bayesian optimization in a billion dimensions via
  random embeddings},
\newblock \bibinfo{journal}{Journal of Artificial Intelligence Research}
  \bibinfo{volume}{55} (\bibinfo{year}{2016}) \bibinfo{pages}{361--387}.
  \DOIprefix\doi{https://doi.org/10.1613/jair.4806}.
\bibitem[{Binois et~al.(2015)Binois, Ginsbourger, and Roustant}]{Binois_2015}
\bibinfo{author}{M.~Binois}, \bibinfo{author}{D.~Ginsbourger},
  \bibinfo{author}{O.~Roustant},
\newblock \bibinfo{title}{A warped kernel improving robustness in {B}ayesian
  optimization via random embeddings},
\newblock in: \bibinfo{booktitle}{International Conference on Learning and
  Intelligent Optimization}, \bibinfo{year}{2015}, pp.
  \bibinfo{pages}{281--286}.
  \DOIprefix\doi{https://doi.org/10.1007/978-3-319-19084-6_28}.
\bibitem[{Nayebi et~al.(2019)Nayebi, Munteanu, and Poloczek}]{Nayebi_2019}
\bibinfo{author}{A.~Nayebi}, \bibinfo{author}{A.~Munteanu},
  \bibinfo{author}{M.~Poloczek},
\newblock \bibinfo{title}{A framework for {B}ayesian optimization in embedded
  subspaces},
\newblock in: \bibinfo{booktitle}{International Conference on Machine
  Learning}, volume~\bibinfo{volume}{97}, \bibinfo{year}{2019}, pp.
  \bibinfo{pages}{4752--4761}.
\bibitem[{Binois et~al.(2020)Binois, Ginsbourger, and Roustant}]{Binois_2020}
\bibinfo{author}{M.~Binois}, \bibinfo{author}{D.~Ginsbourger},
  \bibinfo{author}{O.~Roustant},
\newblock \bibinfo{title}{On the choice of the low-dimensional domain for
  global optimization via random embeddings},
\newblock \bibinfo{journal}{Journal of Global Optimization}
  \bibinfo{volume}{76} (\bibinfo{year}{2020}) \bibinfo{pages}{69--90}.
  \DOIprefix\doi{https://doi.org/10.1007/s10898-019-00839-1}.
\bibitem[{Letham et~al.(2020)Letham, Calandra, Rai, and Bakshy}]{Letham_2020}
\bibinfo{author}{B.~Letham}, \bibinfo{author}{R.~Calandra},
  \bibinfo{author}{A.~Rai}, \bibinfo{author}{E.~Bakshy},
\newblock \bibinfo{title}{Re-examining linear embeddings for high-dimensional
  {B}ayesian optimization},
\newblock in: \bibinfo{booktitle}{Advances in Neural Information Processing
  Systems}, volume~\bibinfo{volume}{33}, \bibinfo{year}{2020}, pp.
  \bibinfo{pages}{1546--1558}.
\bibitem[{Papenmeier et~al.(2022)Papenmeier, Nardi, and
  Poloczek}]{Papenmeier_2022}
\bibinfo{author}{L.~Papenmeier}, \bibinfo{author}{L.~Nardi},
  \bibinfo{author}{M.~Poloczek},
\newblock \bibinfo{title}{Increasing the scope as you learn: Adaptive bayesian
  optimization in nested subspaces},
\newblock in: \bibinfo{booktitle}{Advances in Neural Information Processing
  Systems}, \bibinfo{year}{2022}, pp. \bibinfo{pages}{11586--11601}.
\bibitem[{Qian et~al.(2016)Qian, Hu, and Yu}]{Qian_2016}
\bibinfo{author}{H.~Qian}, \bibinfo{author}{Y.-Q. Hu}, \bibinfo{author}{Y.~Yu},
\newblock \bibinfo{title}{Derivative-free optimization of high-dimensional
  non-convex functions by sequential random embeddings},
\newblock in: \bibinfo{booktitle}{International Joint Conference on Artificial
  Intelligence}, \bibinfo{year}{2016}, pp. \bibinfo{pages}{1946--1952}.
\bibitem[{Bouhlel et~al.(2016)Bouhlel, Bartoli, Otsmane, and
  Morlier}]{Bouhlel_2016}
\bibinfo{author}{M.~A. Bouhlel}, \bibinfo{author}{N.~Bartoli},
  \bibinfo{author}{A.~Otsmane}, \bibinfo{author}{J.~Morlier},
\newblock \bibinfo{title}{Improving {K}riging surrogates of high-dimensional
  design models by partial least squares dimension reduction},
\newblock \bibinfo{journal}{Structural and Multidisciplinary Optimization}
  \bibinfo{volume}{53} (\bibinfo{year}{2016}) \bibinfo{pages}{935--952}.
  \DOIprefix\doi{https://doi.org/10.1007/s00158-015-1395-9}.
\bibitem[{Zhang et~al.(2019)Zhang, Li, and Su}]{Zhang_2019}
\bibinfo{author}{M.~Zhang}, \bibinfo{author}{H.~Li}, \bibinfo{author}{S.~Su},
\newblock \bibinfo{title}{High dimensional {B}ayesian optimization via
  supervised dimension reduction},
\newblock in: \bibinfo{booktitle}{International Joint Conference on Artificial
  Intelligence}, \bibinfo{year}{2019}, pp. \bibinfo{pages}{4292--4298}.
\bibitem[{Moriconi et~al.(2020)Moriconi, Deisenroth, and
  Sesh~Kumar}]{Moriconi_2020b}
\bibinfo{author}{R.~Moriconi}, \bibinfo{author}{M.~P. Deisenroth},
  \bibinfo{author}{K.~S. Sesh~Kumar},
\newblock \bibinfo{title}{High-dimensional bayesian optimization using
  low-dimensional feature spaces},
\newblock \bibinfo{journal}{Machine Learning} \bibinfo{volume}{109}
  (\bibinfo{year}{2020}) \bibinfo{pages}{1925--1943}.
  \DOIprefix\doi{https://doi.org/10.1007/s10994-020-05899-z}.
\bibitem[{Griffiths and Hernández-Lobato(2020)}]{Griffiths_2020}
\bibinfo{author}{R.-R. Griffiths}, \bibinfo{author}{J.~M. Hernández-Lobato},
\newblock \bibinfo{title}{Constrained {B}ayesian optimization for automatic
  chemical design using variational autoencoders},
\newblock \bibinfo{journal}{Chemical science} \bibinfo{volume}{11}
  (\bibinfo{year}{2020}) \bibinfo{pages}{577--586}.
  \DOIprefix\doi{https://doi.org/10.1039/C9SC04026A}.
\bibitem[{Siivola et~al.(2021)Siivola, Paleyes, González, and
  Vehtari}]{Siivola_2021}
\bibinfo{author}{E.~Siivola}, \bibinfo{author}{A.~Paleyes},
  \bibinfo{author}{J.~González}, \bibinfo{author}{A.~Vehtari},
\newblock \bibinfo{title}{Good practices for {B}ayesian optimization of high
  dimensional structured spaces},
\newblock \bibinfo{journal}{Applied AI Letters} \bibinfo{volume}{2}
  (\bibinfo{year}{2021}) \bibinfo{pages}{e24}.
  \DOIprefix\doi{https://doi.org/10.1002/ail2.24}.
\bibitem[{Maus et~al.(2022)Maus, Jones, Moore, Kusner, Bradshaw, and
  Gardner}]{Maus_2022}
\bibinfo{author}{N.~Maus}, \bibinfo{author}{H.~Jones},
  \bibinfo{author}{J.~Moore}, \bibinfo{author}{M.~J. Kusner},
  \bibinfo{author}{J.~Bradshaw}, \bibinfo{author}{J.~Gardner},
\newblock \bibinfo{title}{Local latent space {B}ayesian optimization over
  structured inputs},
\newblock in: \bibinfo{booktitle}{Advances in Neural Information Processing
  Systems}, volume~\bibinfo{volume}{35}, \bibinfo{year}{2022}, pp.
  \bibinfo{pages}{34505--34518}.
\bibitem[{Kandasamy et~al.(2015)Kandasamy, Schneider, and
  Póczos}]{Kandasamy_2015}
\bibinfo{author}{T.~Kandasamy}, \bibinfo{author}{J.~Schneider},
  \bibinfo{author}{B.~Póczos},
\newblock \bibinfo{title}{High dimensional {B}ayesian optimisation and bandits
  via additive models},
\newblock in: \bibinfo{booktitle}{International Conference on Machine
  Learning}, \bibinfo{year}{2015}, p. \bibinfo{pages}{295–304}.
\bibitem[{Li et~al.(2016)Li, Kandasamy, Poczos, and Schneider}]{Li_2016}
\bibinfo{author}{C.-L. Li}, \bibinfo{author}{K.~Kandasamy},
  \bibinfo{author}{B.~Poczos}, \bibinfo{author}{J.~Schneider},
\newblock \bibinfo{title}{High dimensional {B}ayesian optimization via
  restricted projection pursuit models},
\newblock in: \bibinfo{booktitle}{International Conference on Artificial
  Intelligence and Statistics}, volume~\bibinfo{volume}{51},
  \bibinfo{year}{2016}, pp. \bibinfo{pages}{884--892}.
\bibitem[{Rolland et~al.(2018)Rolland, Scarlett, Bogunovic, and
  Cevher}]{Rolland_2018}
\bibinfo{author}{P.~Rolland}, \bibinfo{author}{J.~Scarlett},
  \bibinfo{author}{I.~Bogunovic}, \bibinfo{author}{V.~Cevher},
\newblock \bibinfo{title}{High-dimensional {B}ayesian optimization via additive
  models with overlapping groups},
\newblock in: \bibinfo{booktitle}{International Conference on Artificial
  Intelligence and Statistics}, volume~\bibinfo{volume}{84},
  \bibinfo{year}{2018}, pp. \bibinfo{pages}{298--307}.
\bibitem[{Wang et~al.(2017)Wang, Li, Jegelka, and Kohli}]{Wang_2017}
\bibinfo{author}{Z.~Wang}, \bibinfo{author}{C.~Li},
  \bibinfo{author}{S.~Jegelka}, \bibinfo{author}{P.~Kohli},
\newblock \bibinfo{title}{Batched high-dimensional {B}ayesian optimization via
  structural kernel learning},
\newblock in: \bibinfo{booktitle}{International Conference on Machine
  Learning}, volume~\bibinfo{volume}{70}, \bibinfo{year}{2017}, pp.
  \bibinfo{pages}{3656--3664}.
\bibitem[{Eriksson et~al.(2019)Eriksson, Pearce, Gardner, Turner, and
  Poloczek}]{Eriksson_2019}
\bibinfo{author}{D.~Eriksson}, \bibinfo{author}{M.~Pearce},
  \bibinfo{author}{J.~Gardner}, \bibinfo{author}{R.~D. Turner},
  \bibinfo{author}{M.~Poloczek},
\newblock \bibinfo{title}{Scalable global optimization via local {B}ayesian
  optimization},
\newblock in: \bibinfo{booktitle}{Advances in Neural Information Processing
  Systems}, volume~\bibinfo{volume}{32}, \bibinfo{year}{2019}, pp.
  \bibinfo{pages}{5496--5507}.
\bibitem[{Oh et~al.(2018)Oh, Gavves, and Welling}]{Oh_2018}
\bibinfo{author}{C.~Oh}, \bibinfo{author}{E.~Gavves},
  \bibinfo{author}{M.~Welling},
\newblock \bibinfo{title}{{BOCK} : {B}ayesian optimization with cylindrical
  kernels},
\newblock in: \bibinfo{booktitle}{International Conference on Machine
  Learning}, volume~\bibinfo{volume}{80}, \bibinfo{year}{2018}, pp.
  \bibinfo{pages}{3868--3877}.
\bibitem[{Kirschner et~al.(2019)Kirschner, Mutny, Hiller, Ischebeck, and
  Krause}]{Kirschner_2019}
\bibinfo{author}{J.~Kirschner}, \bibinfo{author}{M.~Mutny},
  \bibinfo{author}{N.~Hiller}, \bibinfo{author}{R.~Ischebeck},
  \bibinfo{author}{A.~Krause},
\newblock \bibinfo{title}{Adaptive and safe {B}ayesian optimization in high
  dimensions via one-dimensional subspaces},
\newblock in: \bibinfo{booktitle}{International Conference on Machine
  Learning}, volume~\bibinfo{volume}{97}, \bibinfo{year}{2019}, pp.
  \bibinfo{pages}{3429--3438}.
\bibitem[{Frazier(2018)}]{Frazier_2018}
\bibinfo{author}{P.~I. Frazier},
\newblock \bibinfo{title}{A tutorial on {B}ayesian optimization},
\newblock \bibinfo{journal}{arXiv:.02811}  (\bibinfo{year}{2018}).
  \DOIprefix\doi{https://doi.org/10.48550/arXiv.1807.02811}.
\bibitem[{Wang et~al.(2023)Wang, Jin, Schmitt, and Olhofer}]{Wang_2023}
\bibinfo{author}{X.~Wang}, \bibinfo{author}{Y.~Jin},
  \bibinfo{author}{S.~Schmitt}, \bibinfo{author}{M.~Olhofer},
\newblock \bibinfo{title}{Recent advances in {B}ayesian optimization},
\newblock \bibinfo{journal}{ACM Computing Surveys} \bibinfo{volume}{55}
  (\bibinfo{year}{2023}) \bibinfo{pages}{Article 287}.
  \DOIprefix\doi{https://doi.org/10.1145/3582078}.
\bibitem[{Forrester and Keane(2008)}]{Forrester_2008}
\bibinfo{author}{A.~Forrester}, \bibinfo{author}{A.~Keane},
  \bibinfo{title}{Engineering design via surrogate modelling: a practical
  guide}, \bibinfo{publisher}{John Wiley \& Sons}, \bibinfo{year}{2008}.
\bibitem[{Frazier et~al.(2008)Frazier, Powell, and Dayanik}]{Frazier_2008}
\bibinfo{author}{P.~I. Frazier}, \bibinfo{author}{W.~B. Powell},
  \bibinfo{author}{S.~Dayanik},
\newblock \bibinfo{title}{A knowledge-gradient policy for sequential
  information collection},
\newblock \bibinfo{journal}{SIAM Journal on Control and Optimization}
  \bibinfo{volume}{47} (\bibinfo{year}{2008}) \bibinfo{pages}{2410--2439}.
  \DOIprefix\doi{https://doi.org/10.1137/070693424}.
\bibitem[{Frazier et~al.(2009)Frazier, Powell, and Dayanik}]{Frazier_2009}
\bibinfo{author}{P.~Frazier}, \bibinfo{author}{W.~Powell},
  \bibinfo{author}{S.~Dayanik},
\newblock \bibinfo{title}{The knowledge-gradient policy for correlated normal
  beliefs},
\newblock \bibinfo{journal}{INFORMS Journal on Computing} \bibinfo{volume}{21}
  (\bibinfo{year}{2009}) \bibinfo{pages}{599--613}.
  \DOIprefix\doi{https://doi.org/10.1287/ijoc.1080.0314}.
\bibitem[{Hennig and Schuler(2012)}]{Hennig_2012}
\bibinfo{author}{P.~Hennig}, \bibinfo{author}{C.~J. Schuler},
\newblock \bibinfo{title}{Entropy search for information-efficient global
  optimization},
\newblock \bibinfo{journal}{Journal of Machine Learning Research}
  \bibinfo{volume}{13} (\bibinfo{year}{2012}) \bibinfo{pages}{1809--1837}.
\bibitem[{Hern\'{a}ndez-Lobato et~al.(2014)Hern\'{a}ndez-Lobato, Hoffman, and
  Ghahramani}]{Hernandez_2014}
\bibinfo{author}{J.~M. Hern\'{a}ndez-Lobato}, \bibinfo{author}{M.~W. Hoffman},
  \bibinfo{author}{Z.~Ghahramani},
\newblock \bibinfo{title}{Predictive entropy search for efficient global
  optimization of black-box functions},
\newblock in: \bibinfo{booktitle}{Advances in Neural Information Processing
  Systems}, \bibinfo{year}{2014}, pp. \bibinfo{pages}{918--926}.
\bibitem[{Zhan and Xing(2020)}]{Zhan_2020}
\bibinfo{author}{D.~Zhan}, \bibinfo{author}{H.~Xing},
\newblock \bibinfo{title}{Expected improvement for expensive optimization: a
  review},
\newblock \bibinfo{journal}{Journal of Global Optimization}
  \bibinfo{volume}{78} (\bibinfo{year}{2020}) \bibinfo{pages}{507--544}.
  \DOIprefix\doi{https://doi.org/10.1007/s10898-020-00923-x}.
\bibitem[{Liang et~al.(2013)Liang, Qu, Suganthan, and
  Hernández-Díaz}]{Liang_2013}
\bibinfo{author}{J.~J. Liang}, \bibinfo{author}{B.~Qu}, \bibinfo{author}{P.~N.
  Suganthan}, \bibinfo{author}{A.~G. Hernández-Díaz}, \bibinfo{title}{Problem
  definitions and evaluation criteria for the CEC 2013 special session on
  real-parameter optimization}, \bibinfo{type}{Technical Report}, Computational
  Intelligence Laboratory, Zhengzhou University, Zhengzhou, China and Nanyang
  Technological University, Singapore, \bibinfo{year}{2013}.
\bibitem[{Awad et~al.(2017)Awad, Ali, Liang, Qu, and Suganthan}]{Awad_2017}
\bibinfo{author}{N.~Awad}, \bibinfo{author}{M.~Ali},
  \bibinfo{author}{J.~Liang}, \bibinfo{author}{B.~Qu},
  \bibinfo{author}{P.~Suganthan}, \bibinfo{title}{Problem definitions and
  evaluation criteria for the {CEC} special session and competition on single
  objective real-parameter numerical optimization}, \bibinfo{type}{Technical
  Report}, Nanyang Technological University, Singapore, Jordan University of
  Science and Technology, Jordan and Zhengzhou University, Zhengzhou China,
  \bibinfo{year}{2017}.
\bibitem[{Xie et~al.(2023)Xie, Li, Wang, Cui, and Gong}]{Xie_2023}
\bibinfo{author}{L.~Xie}, \bibinfo{author}{G.~Li}, \bibinfo{author}{Z.~Wang},
  \bibinfo{author}{L.~Cui}, \bibinfo{author}{M.~Gong},
\newblock \bibinfo{title}{Surrogate-assisted evolutionary algorithm with model
  and infill criterion auto-configuration},
\newblock \bibinfo{journal}{IEEE Transactions on Evolutionary Computation}
  (\bibinfo{year}{2023}) \bibinfo{pages}{1--1}.
  \DOIprefix\doi{10.1109/TEVC.2023.3291614}.
\bibitem[{Li et~al.(2023{\natexlab{a}})Li, Xie, Wang, Wang, and
  Gong}]{LiGH_2023}
\bibinfo{author}{G.~Li}, \bibinfo{author}{L.~Xie}, \bibinfo{author}{Z.~Wang},
  \bibinfo{author}{H.~Wang}, \bibinfo{author}{M.~Gong},
\newblock \bibinfo{title}{Evolutionary algorithm with individual-distribution
  search strategy and regression-classification surrogates for expensive
  optimization},
\newblock \bibinfo{journal}{Information Sciences} \bibinfo{volume}{634}
  (\bibinfo{year}{2023}{\natexlab{a}}) \bibinfo{pages}{423--442}.
  \DOIprefix\doi{https://doi.org/10.1016/j.ins.2023.03.101}.
\bibitem[{Li et~al.(2023{\natexlab{b}})Li, Wang, and Gong}]{LiG_2023}
\bibinfo{author}{G.~Li}, \bibinfo{author}{Z.~Wang}, \bibinfo{author}{M.~Gong},
\newblock \bibinfo{title}{Expensive optimization via surrogate-assisted and
  model-free evolutionary optimization},
\newblock \bibinfo{journal}{IEEE Transactions on Systems, Man, and Cybernetics:
  Systems} \bibinfo{volume}{53} (\bibinfo{year}{2023}{\natexlab{b}})
  \bibinfo{pages}{2758--2769}. \DOIprefix\doi{10.1109/TSMC.2022.3219080}.
\bibitem[{Liu et~al.(2014)Liu, Zhang, and Gielen}]{Liu_2014}
\bibinfo{author}{B.~Liu}, \bibinfo{author}{Q.~Zhang}, \bibinfo{author}{G.~G.~E.
  Gielen},
\newblock \bibinfo{title}{A gaussian process surrogate model assisted
  evolutionary algorithm for medium scale expensive optimization problems},
\newblock \bibinfo{journal}{IEEE Transactions on Evolutionary Computation}
  \bibinfo{volume}{18} (\bibinfo{year}{2014}) \bibinfo{pages}{180--192}.
  \DOIprefix\doi{10.1109/TEVC.2013.2248012}.
\bibitem[{Tian et~al.(2019)Tian, Tan, Zeng, Sun, and Jin}]{Tian_2019}
\bibinfo{author}{J.~Tian}, \bibinfo{author}{Y.~Tan}, \bibinfo{author}{J.~Zeng},
  \bibinfo{author}{C.~Sun}, \bibinfo{author}{Y.~Jin},
\newblock \bibinfo{title}{Multiobjective infill criterion driven gaussian
  process-assisted particle swarm optimization of high-dimensional expensive
  problems},
\newblock \bibinfo{journal}{IEEE Transactions on Evolutionary Computation}
  \bibinfo{volume}{23} (\bibinfo{year}{2019}) \bibinfo{pages}{459--472}.
  \DOIprefix\doi{10.1109/TEVC.2018.2869247}.
\bibitem[{Zhan and Xing(2021)}]{Zhan_2021}
\bibinfo{author}{D.~Zhan}, \bibinfo{author}{H.~Xing},
\newblock \bibinfo{title}{A fast kriging-assisted evolutionary algorithm based
  on incremental learning},
\newblock \bibinfo{journal}{IEEE Transactions on Evolutionary Computation}
  \bibinfo{volume}{25} (\bibinfo{year}{2021}) \bibinfo{pages}{941--955}.
  \DOIprefix\doi{10.1109/TEVC.2021.3067015}.
\bibitem[{Zhan et~al.(2023)Zhan, Gui, and Li}]{Zhan_2023b}
\bibinfo{author}{D.~Zhan}, \bibinfo{author}{Y.~Gui}, \bibinfo{author}{T.~Li},
\newblock \bibinfo{title}{An anisotropic expected improvement criterion for
  kriging-assisted evolutionary computation},
\newblock in: \bibinfo{booktitle}{IEEE Congress on Evolutionary Computation
  (CEC)}, \bibinfo{year}{2023}, pp. \bibinfo{pages}{1--8}.
  \DOIprefix\doi{10.1109/CEC53210.2023.10254097}.
\bibitem[{Si et~al.(2023)Si, Zhang, Tian, Yang, Zhang, and Jin}]{Si_2023}
\bibinfo{author}{L.~Si}, \bibinfo{author}{X.~Zhang}, \bibinfo{author}{Y.~Tian},
  \bibinfo{author}{S.~Yang}, \bibinfo{author}{L.~Zhang},
  \bibinfo{author}{Y.~Jin},
\newblock \bibinfo{title}{Linear subspace surrogate modeling for large-scale
  expensive single/multi-objective optimization},
\newblock \bibinfo{journal}{IEEE Transactions on Evolutionary Computation}
  (\bibinfo{year}{2023}) \bibinfo{pages}{1--16}.
  \DOIprefix\doi{10.1109/TEVC.2023.3319640}.
\bibitem[{Deb and Jain(2014)}]{Deb_2014}
\bibinfo{author}{K.~Deb}, \bibinfo{author}{H.~Jain},
\newblock \bibinfo{title}{An evolutionary many-objective optimization algorithm
  using reference-point-based nondominated sorting approach, part i: Solving
  problems with box constraints},
\newblock \bibinfo{journal}{IEEE Transactions on Evolutionary Computation}
  \bibinfo{volume}{18} (\bibinfo{year}{2014}) \bibinfo{pages}{577--601}.
  \DOIprefix\doi{10.1109/TEVC.2013.2281535}.
\bibitem[{Tian et~al.(2017)Tian, Cheng, Zhang, and Jin}]{Tian_2017}
\bibinfo{author}{Y.~Tian}, \bibinfo{author}{R.~Cheng},
  \bibinfo{author}{X.~Zhang}, \bibinfo{author}{Y.~Jin},
\newblock \bibinfo{title}{Platemo: A matlab platform for evolutionary
  multi-objective optimization [educational forum]},
\newblock \bibinfo{journal}{IEEE Computational Intelligence Magazine}
  \bibinfo{volume}{12} (\bibinfo{year}{2017}) \bibinfo{pages}{73--87}.
  \DOIprefix\doi{10.1109/MCI.2017.2742868}.

\end{thebibliography}



\end{document}